\documentclass{article}

    \PassOptionsToPackage{numbers,sort&compress}{natbib}

\usepackage[preprint]{neurips_2025}
\makeatletter
\renewcommand{\@noticestring}{}
\makeatother

\usepackage[utf8]{inputenc} 
\usepackage[T1]{fontenc}    
\usepackage{hyperref}       
\usepackage{url}            
\usepackage{booktabs}       
\usepackage{amsfonts}       
\usepackage{nicefrac}       
\usepackage{microtype}      
\usepackage{xcolor}         

\usepackage[pdftex]{graphicx}
\graphicspath{ {./figures/} }
\usepackage[dvipsnames]{xcolor}
\usepackage{amsmath}
\usepackage{pifont}
\newcommand{\cmark}{\ding{51}}  
\newcommand{\xmark}{\ding{55}}  
\usepackage{amsthm}
\theoremstyle{plain}
\newtheorem{theorem}{Theorem}[section]
\newtheorem{proposition}[theorem]{Proposition}

\title{Softmax $\geq$ Linear: Transformers may learn to classify in-context by kernel gradient descent}

\author{
  Sara Dragutinović$^{1}$ \quad
  Andrew M. Saxe$^{2,*}$ \quad
  Aaditya K. Singh$^{2,*}$ \\[0.5em]
  $^{1}$Department of Computer Science, University of Oxford \\
  $^{2}$Gatsby Computational Neuroscience Unit, University College London
}

\newcommand{\R}{\mathbb{R}}
\newcommand{\bz}{\bold{z}}
\newcommand{\bx}{\bold{x}}
\newcommand{\by}{\bold{y}}
\newcommand{\bw}{\bold{w}}
\newcommand{\ba}{\bold{a}}
\newcommand{\bb}{\bold{b}}

\newcommand{\q}{\text{query}}
\newcommand{\softmax}{\text{softmax}}
\newcommand{\sfmx}{\text{sfmx}}

\begin{document}

\maketitle

\begin{abstract}
  The remarkable ability of transformers to learn new concepts solely by reading examples within the input prompt, termed in-context learning (ICL), is a crucial aspect of intelligent behavior. Here, we focus on understanding the learning algorithm transformers use to learn from context. Existing theoretical work, often based on simplifying assumptions, has primarily focused on linear self-attention and continuous regression tasks, finding transformers can learn in-context by gradient descent. Given that transformers are typically trained on discrete and complex tasks, we bridge the gap from this existing work to the setting of \textit{classification}, with \textit{non-linear} (importantly, \textit{softmax}) activation. We find that transformers still learn to do gradient descent in-context, though on functionals in the kernel feature space and with a context-adaptive learning rate in the case of softmax transformer. These theoretical findings suggest a greater adaptability to context for softmax attention, which we empirically verify and study through ablations. Overall, we hope this enhances theoretical understanding of in-context learning algorithms in more realistic settings, pushes forward our intuitions and enables further theory bridging to larger models.\footnote{Correspondence to: \texttt{sara.dragutinovic@cs.ox.ac.uk}. *Equal senior authorship. Code available at \href{https://github.com/saradrag/transformer-icl-classification}{github.com/saradrag/transformer-icl-classification}.}
\end{abstract}

\section{Introduction}
\label{sec:intro}
With the rapid advancement of machine learning architectures, a new wave of questions has emerged---not only about achieving state-of-the-art performance, but also about understanding \textit{how} models solve tasks. As models grow larger---often containing tens of billions of parameters---the trend of scaling has revealed surprising emergent behaviors. Among these is in-context learning (ICL), a capability that has prompted increasing interest and investigation.

\begin{figure}
    \centering
    \includegraphics[width=0.73\linewidth]{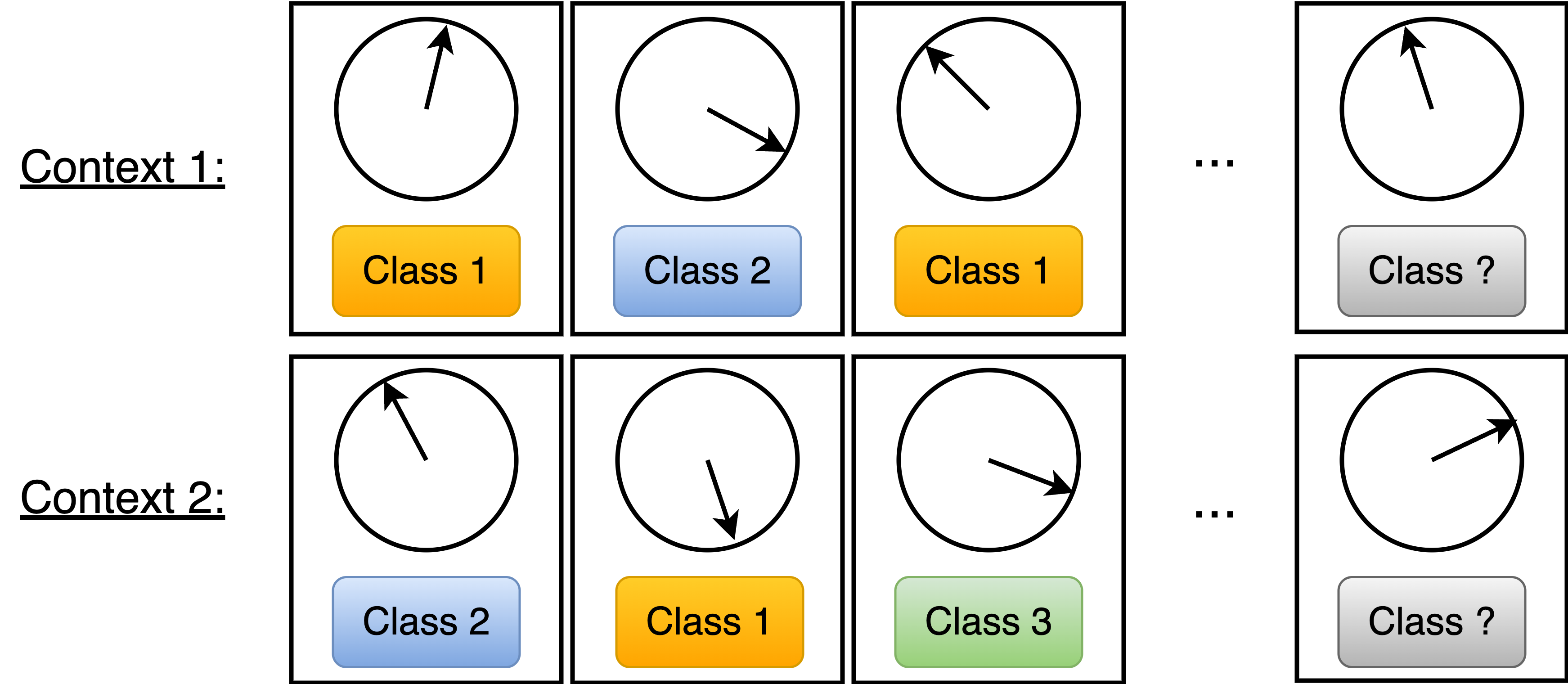}
    \caption{Two example contexts from our classification task (see Section~\ref{sec:task_setup}). For Context 1, the correct class for query is Class 1, as $\bx_\q$ lays between the first and the third context vector. With Context 2, we emphasize how: 1) context vectors $\bx_i$ and query input $\bx_\q$ differ between contexts; 2) class assignment differs between contexts---if the first context vector was in Context 1, its label would be Class 1, not Class 2.}
    \label{fig:contexts}
\end{figure}

ICL is a critical ability often exhibited by transformers whereby they can adapt to context (i.e. input prompt) and solve tasks not present during training. After its initial emergence when training transformers at scale on natural language \cite{brown2020language}, much empirical and theoretical work has gone into studying this phenomena. In a controlled setting, experiments on ICL typically involve a prompt containing several input-label (or question-answer) pairs, followed by a query input for which the model is expected to produce the correct output, as depicted in Figure~\ref{fig:contexts}. Following prior work \cite{garg2022can, akyurek2022learning, von2023transformers, ahn2023transformers}, this paper aims to contribute to answering the question: ‘What learning algorithms do transformers use to perform ICL?’

Prior theoretical work suggests that transformers may implement in-context learning by taking a step of (preconditioned) gradient descent (GD) \cite{von2023transformers, garg2022can, akyurek2022learning, ahn2023transformers, mahankali2023one}. However, much of this work has been restricted to linear regression tasks and transformers with a linear activation function, both of which are departures from standard settings \cite{vaswani2017attention}. Here, we emphasize the key departures from prior work, and how they are bridging the gaps between theoretical results and the transformers used in practice:
\begin{enumerate}
    \item \textbf{Switching to classification task}: Learning a continuous task with mean squared error (MSE) loss and learning a discrete task with cross-entropy (CE) loss are fundamentally different. Minimizing MSE corresponds to maximizing likelihood \textit{under a Gaussian noise assumption}, whereas minimizing CE loss requires only the assumption of i.i.d. samples, without any additional noise model. From a practical standpoint, many transformers are trained to predict a probability distribution over a discrete set of tokens---most notably, large language models (LLMs) predicting the next token in a sequence—making the classification setup more representative of real-world applications.
    \item \textbf{Using softmax attention}: From the original introduction of transformers \cite{vaswani2017attention} to today's LLMs and vision transformers, the softmax activation in self-attention has remained the standard choice. On the other hand, the mathematical convenience of linear attention (i.e., attention without any activation function) has attracted much of the existing theoretical analysis. Motivated by the intuition that linear and softmax attention differ not only in implementation but potentially in behavior, we aim to understand both in the ICL setting, with a particular focus on how the learning algorithms they implement may differ.
\end{enumerate}
Specifically, we find that self-attention can be seen as a step of gradient descent even when doing a discrete task, on CE loss. While linear self-attention still implements the usual step of GD on context data, softmax self-attention implements a step of functional gradient descent in the Radial Basis Function (RBF) kernel feature space (kernel GD); furthermore, it benefits from having a context-adaptive learning rate. With two effective parameters (kernel width and learning rate), softmax attention outperforms linear attention (with learning rate only) on the challenging setup of our synthetic ICL classification task. Investigating this performance gap, we build intuition on what it means to do gradient descent on an input context (where it works well, and where it possibly doesn’t). Overall, our results point to how seemingly simple design choices can impact theoretical conclusions, and how these conclusions can lead to qualitatively different behaviors.

The main contributions can be summarized as follows:
\begin{itemize}
    \item We provide weights constructions for \textcolor{BrickRed}{linear}, \textcolor{OliveGreen}{kernel activated} and \textcolor{NavyBlue}{softmax} self-attention, under which they implement \textcolor{BrickRed}{one step of GD} [Section~\ref{sec:theory_lin}], \textcolor{OliveGreen}{a step of kernel GD} [Section~\ref{sec:theory_kernel}] and \textcolor{NavyBlue}{a context-adaptive step of kernel GD} [Section~\ref{sec:theory_sfmx}], respectively, in a \textit{classification} setup with \textit{cross-entropy} loss;
    \item We empirically validate that linear and softmax self-attention---trained to solve a synthetic ICL classification task---both make use of the theoretical solutions we constructed [Section~\ref{sec:experiments}];
    \item We investigate the performance gap between linear and softmax self-attention in challenging settings and conclude that both the correct kernel shape and the context-adaptive learning rate are crucial components of softmax attention success [Section~\ref{sec:softmax>linear}].
\end{itemize}
\section{Setup}
\label{sec:setup}
In this section we introduce the in-context learning task we focus on and the one layer transformer models we use. We emphasize two key choices in our setup---introducing a classification problem and using self-attention with non-linearities (especially, softmax)---as important steps toward bringing prior theoretical analyses closer to practical use cases.
\subsection{Classification Task}
\label{sec:task_setup}
 Following the usual, controlled setup in theoretical work on ICL, the model is provided with a task context $\{(\bold x_i, y_i)\}_{i=1}^n$ (where $n$ is the number of samples in the context), together with the query token $\bold x_{\q}$, and is expected to output $y_\q$, the label of $\bold x_\q$.
 
\begin{figure}
    \centering
    \includegraphics[width=0.71\linewidth]{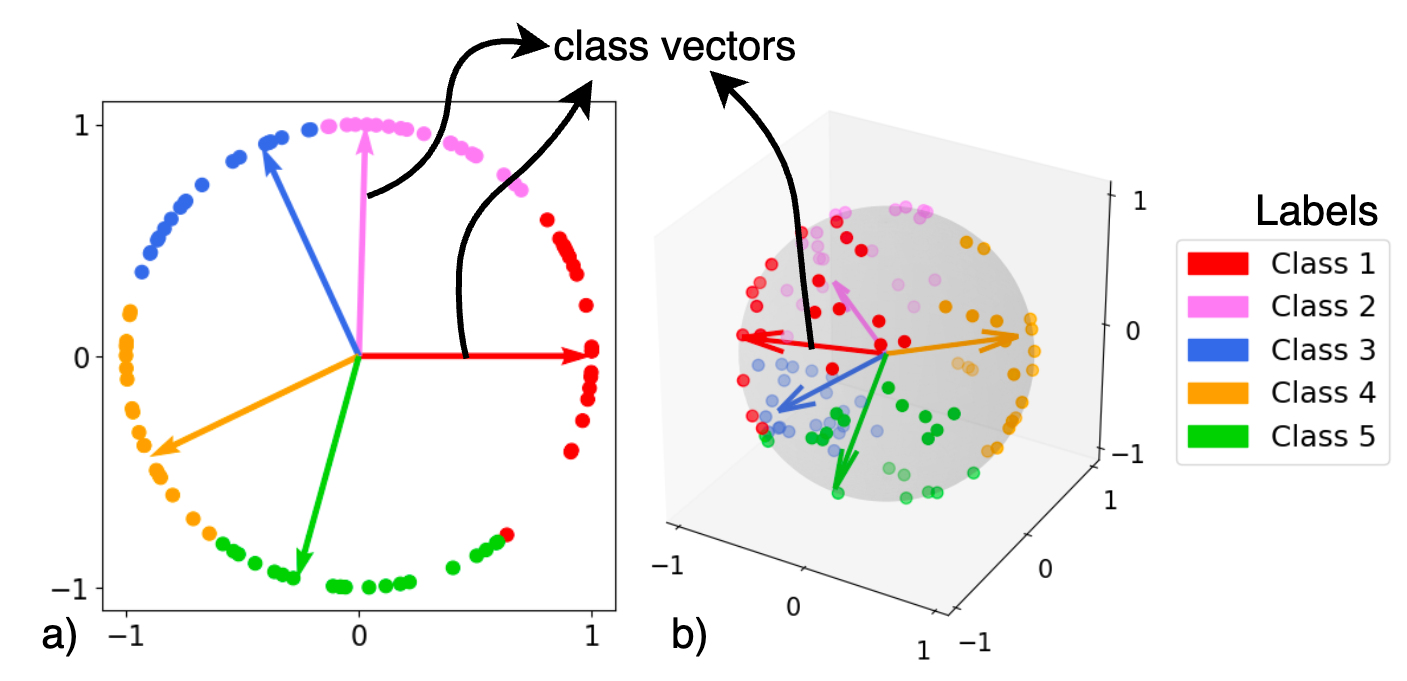}
    \caption{Two different contexts of our synthetic classification task, with $C=5, n=100$ and a)~$d=2$, b)~$d=3$. The arrows are representing class vectors, and the points context vectors; different colors correspond to different class labels.}
    \label{fig:task_examples}
\end{figure}

In our classification setup, the task context is generated as follows. We draw $C$ ``prototype'' class vectors $\bz_1,...,\bz_C\sim U(S^{d-1})$ independently and uniformly from the unit sphere $S^{d-1}$ of $(\R^d, \ell^2)$. Class vectors determine the way we assign labels, and are \textit{not} provided explicitly to the model. Context vectors $\bx$ are also on the unit sphere, and are assigned the class of the closest class vector. We uniformly draw $\frac n C$ context vectors $\bx_i\in S^{d-1}$ from each of the $C$ classes (using rejection sampling),\footnote{We make this choice to avoid models learning pathological strategies such as picking the most common label from context.} with $y_i$ being the corresponding class labels. Lastly, we draw a random query token $\bx_\q$ and its label $y_\q$ from the same distribution as the context vectors: first, a class $y_\q$ is sampled uniformly from all classes, and then $\bx_\q$ is sampled uniformly from the region of the unit sphere $S^{d-1}$ assigned to the class $y_\q$.
The illustrations of the task for $d=2$ and $d=3$ are shown in Figure~\ref{fig:task_examples}. To avoid any confusion, we provide an example of two different contexts in Figure~\ref{fig:contexts}. While seemingly simple, this discrete task is highly non-linear, enabling us to study transformers in a regime closer to practice.

\subsection{Model Setup}
\label{sec:model_setup}
\begin{figure}
    \centering
    \includegraphics[width=\linewidth]{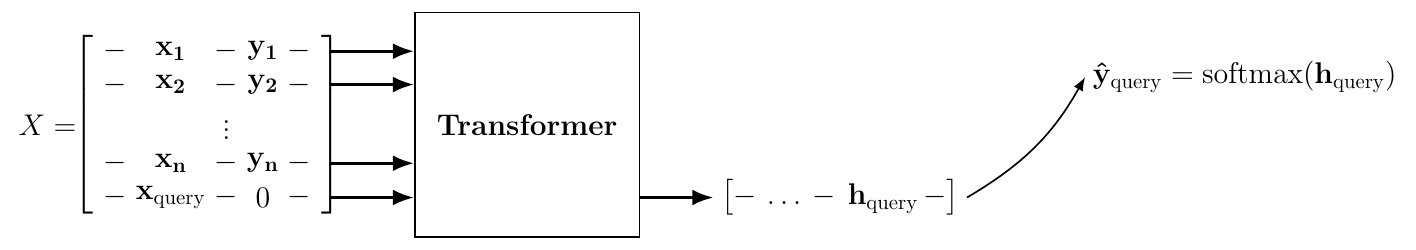}
    \caption{Illustration of a transformer forward pass. In our setting, we only make use of the update of the last token---we extract the $y$-entry of it and apply softmax to get the final prediction $\hat \by_\q$.}
    \label{fig:fwd_pass}
\end{figure}

Following the setup of \citep{von2023transformers}, we use a simplified one layer transformer with a single head. As common in prior work \cite{huang2023context, cheng2023transformers, li2024nonlinear, zhang2025training}, we use concatenated input-output pairs as tokens: $[\bx_i, \by_i]\in \R^{d+C}$, where $\by_i$ is one-hot encoding of $y_i$. The query token is concatenated with $\bold 0\in \R^C$ vector as its label. Input to the transformer is a matrix $X$ whose rows are all such tokens, as seen in Figure~\ref{fig:fwd_pass}.

Suppose $W_Q, W_K, W_V$ and $W_O$ are the $(d+C)\times (d+C)$ query, key, value and projection/output weight matrices. The formula for (unmasked\footnote{The masking actually doesn't matter in our case, because we're predicting the label of $\bx_\q$ from the last token, which can attend to all previous tokens either with or without the causal masking.}) self-attention (SA) forward pass is
\begin{equation*}
    SA(X)= X + f\left(XW_Q^\top W_KX^\top \right)XW_V^\top W_O^\top 
\end{equation*}
where $f$ is the activation function applied to attention matrix. We theoretically analyze three different models, differing in $f$: linear self-attention, kernel self-attention, and softmax self-attention.

\textbf{Linear self-attention.} In the case of linear SA, $f$ is the identity map. Given $\bx_\q$, linear SA makes the following prediction for $\by_\q$:
\[\hat \by_\q = \softmax\bigg \{\left[\left([\bx_\q, \bold 0]W_Q^\top W_KX^\top  \right)XW_V^\top W_O^\top \right]_y\bigg \}.\]Note that the $\softmax$ in the formula has nothing to do with activation on the attention; it is used as we're training on a classification task, and matches what is commonly done in practice with cross-entropy loss.
The logits of our prediction for $\by_\q$ is the $y$-entry of the last, query token (as seen in Figure~\ref{fig:fwd_pass}). We denote this by subscript $y$ in the formula. 

\textbf{Kernel self-attention.}
In the case of kernel self-attention, the activation function on the attention originates from a kernel $k:\R^{d+C}\times\R^{d+C}\to \R$ (symmetric and positive semi-definite). Specifically, those activation functions $f=\text{act}_k$ satisfy the following: for two matrices $A = [\ba_1,...\ba_a], B=[\bb_1,...,\bb_b]$, applying kernel activation gives a $a\times b$-sized matrix ${\text{act}_k([\ba_1,...,\ba_a], [\bb_1,...,\bb_b]) = \{k(\ba_i, \bb_j)\}_{i=1,...,a;j=1,...,b}}$. With this, the prediction for $\by_\q$ is
\[\hat \by_\q = \softmax\bigg \{\left[\text{act}_k\left(W_Q[\bx_\q, \bold 0]^\top , W_KX^\top  \right)XW_V^\top W_O^\top \right]_y\bigg \}.\]
Note that using $k(\bx, \bx') = \bx^\top \bx'$ is equivalent to linear self-attention.

\textbf{Softmax self-attention.} The most commonly used activation function is $f=\softmax$, coming from the original architecture \cite{von2023transformers}, making the prediction of interest be
\[\hat \by_\q = \softmax\left \{\left[\softmax\left(\frac {[\bx_\q, \bold 0]W_Q^\top W_KX^\top } {\sqrt{d+C}}\right)XW_V^\top W_O^\top \right]_y\right \}.\]
As per usual, we divide by the normalization constant $\sqrt{d+C}$, where $d+C$ is the dimension of the model. Softmax self-attention is \textit{not} a form of kernel SA, due to the normalization inherent to the softmax function.

\section{Transformers \textit{can} implement kernel gradient descent on classification tasks}
\label{sec:theory}
To build up to our main theoretical result, namely that \textit{softmax} transformers can implement \textit{adaptive learning rate kernel gradient descent}, we iteratively relax assumptions made in prior work \cite{von2023transformers}, which showed that linear transformers can implement GD on linear regression tasks.

\subsection{Linear transformer on linear classification task}
\label{sec:theory_lin}
First, we bridge the gap from regression tasks with MSE loss to classification tasks with CE loss.

\textbf{Linear classification task:} For consistency, here we introduce what we call \textit{linear classification}, and is also referred to by names softmax regression or multinomial logistic regression. We are given data $\{(\bx_i, \by_i)\}_{i=1}^n$, where $\by_i\in \R^C$ are one-hot encoded labels. For a new data point $\bx_\q\in \R^d$, we give prediction ${\mathbb P[\bx_\q\text{ is in class }j]=(\softmax(W^\top \bx_\q))_j}$ (for $j\in \{1,...,C\}$), where $W\in \R^{d\times C}$ are the parameters we optimize for.
\begin{proposition}\label{prop1}
    Linear self-attention is expressive enough to implement one step of gradient descent on cross-entropy (CE) loss in the linear classification setup, assuming we start from $W_0=\bold 0$.\footnote{Note that the assumption of starting from $W_0=\bold 0$ corresponds to no prior knowledge on the classes, a realistic assumption in this setting that is similar to those made by prior works \cite{von2023transformers}.}
\end{proposition} 

\textit{Proof sketch.} Cross-entropy loss and its gradient on one sample $(\bx, \by)$ are given by formulas
\begin{equation*}
    L(W) =- \by^\top \log \softmax(\bz); ~~~~~~~\nabla_WL = \bx(\nabla_{\bz}L)^\top = \bx(\softmax(\bz)-\by)^\top,\\
\end{equation*}
where $\bz=W^\top \bx$.
This means one step of gradient descent on the whole dataset gives:
\begin{equation*}
\begin{split}
    W_{\text{new}} = W_0 - \frac \eta n \sum_{i=1}^n \bx_i(\softmax(W_0^\top \bx_i) - \by_i)^\top ,
\end{split}
\end{equation*}
and assuming $W_0=\bold 0$, this makes a new prediction on $\bx_\q$:
\begin{align}\nonumber
    \hat \by_\q &= \softmax\{W_{\text{new}}^\top \bx_\q\}\\\nonumber
    & = \softmax\bigg \{- \frac \eta n \sum_{i=1}^n \left(\frac 1 C\bold 1-\by_i\right)\bx_i^\top \bx_\q\bigg \}\\
    & = \softmax\bigg \{ \frac \eta n \sum_{i=1}^n \by_i\bx_i^\top \bx_\q\bigg \},\label{eq1}
\end{align}
where crucially we use the fact that softmax is shift invariant. For more details on the derivation and a set of transformer weights implementing this update, see Appendix~\ref{sec:A lin}.
Intuitively, Equation \ref{eq1} can be implemented with linear self-attention because $\bx_i^\top \bx_\q$ can be obtained in the product of key and query matrices, and multiplier $\by_i$ can be extracted using the value matrix. $\Box$

\subsection{Any kernel activation transformer on linear
classification task}
\label{sec:theory_kernel}
Next, we consider a kernel activation transformer on a classification task, to bridge the gap between linear and softmax attention. We build upon \cite{cheng2023transformers} to show that kernel activation self-attention can implement one step of gradient descent in the kernel feature space on this \textit{classification} task.
\begin{proposition}\label{prop2}
    Kernel activation self-attention is expressive enough to implement one step of gradient descent in the kernel feature space, on cross-entropy (CE) loss in the linear classification setup, assuming we start from $W_0=\bold 0$.\footnote{For infinite dimensional kernel feature spaces, what we mean by $W_0 = \bold 0$ is that we're starting a step of functional GD from the zero functional (see Appendix~\ref{sec: A ker theory}).} 
\end{proposition} 
\textit{Proof sketch.} We provide intuition here for finite dimensional kernel feature spaces, with a fully general proof (extending to functionals in RKHS) in Appendix~\ref{sec: A ker theory}.

Let $k$ be a kernel with finite dimensional kernel space and $\varphi$ be its feature map, i.e. $k(\bx, \bx') = \langle \varphi(\bx), \varphi(\bx')\rangle$. 
Using kernel feature expansion, on input $\bx$, we predict $\hat \by = \softmax(W^\top \varphi(\bx))$. All the equations from Section~\ref{sec:theory_lin} hold through, with $\bx$ replaced by $\varphi(\bx)$. Hence we end up with 
\begin{equation}
    \hat \by_\q =\softmax\bigg \{ \frac \eta n \sum_{i=1}^n \by_i\varphi(\bx_i)^\top \varphi(\bx_\q)\bigg \} = \softmax\bigg \{ \frac \eta n \sum_{i=1}^n \by_i k(\bx_i,\bx_\q)\bigg \}. \label{eq2}
\end{equation}
Similarly to the linear transformer case, Equation~\ref{eq2} can be implemented with a kernel activation SA, as $k(\bx_i, \bx_\q)$ can be obtained from the attention, and $\by_i$ can be extracted using the value matrix. $\Box$

Equation \ref{eq2} holds for any kernel, even if the kernel feature space is infinite-dimensional, see the full proof in Appendix~\ref{sec: A ker theory}. The weights implementing it can be found in Appendix~\ref{sec:A ker}.
\subsection{Softmax transformer on linear classification task}
\label{sec:theory_sfmx}
Finally, we show that softmax self-attention implements a step of kernel gradient descent, using the RBF kernel, but with a \textit{context-adaptive learning rate}. We assume that all input vectors have norm one,\footnote{Such an assumption is reasonable given the use of LayerNorm \cite{ba2016layer} in most transformers.} that is $\|\bx_i\|=1$. Starting from Equation \ref{eq2}, and substituting the RBF kernel ${k(\bx, \bx') = e^{-\frac {\|\bx-\bx'\|^2} {2\sigma^2}}}$ with kernel width $\sigma$, we have that one step of kernel gradient descent gives
\begin{align*}
    \hat \by_\q &= \softmax\bigg \{ \frac \eta n \sum_{i=1}^n \by_i e^{-\frac {\|\bx_i-\bx_\q\|^2}{2\sigma^2}}\bigg \}\\
    &=\softmax\bigg \{ \frac \eta n \sum_{i=1}^n \by_i e^{\frac {2\bx_i^\top \bx_\q-2}{2\sigma^2}}\bigg \}\\
    &=\softmax\bigg \{ \frac \eta {ne^{\frac 1 {\sigma^2}}} \sum_{i=1}^n \by_i e^{\frac {\bx_i^\top \bx_\q}{\sigma^2}}\bigg \}.
\end{align*}
If we let $\sigma^2=\frac {\sqrt {d+C}} {c_{\sigma}}$ and define an adaptive learning rate (as a function of $X$ - sample points): ${\eta(X) = \frac {c_{\eta}e^{1/\sigma^2}n} {\sum_{i=1}^ne^{\bx_i^\top \bx_\q/\sigma^2}}}$, for some constants $c_{\sigma}, c_{\eta}>0$, we get
\begin{align}
    \hat \by_\q &=\softmax\left \{ c_{\eta}\sum_{i=1}^n \by_i \frac{e^{\frac {c_{\sigma}\bx_i^\top \bx_\q}{\sqrt {d+C}}}}{\sum_{i=1}^ne^{\frac {c_{\sigma}\bx_i^\top \bx_\q}{\sqrt {d+C}}}}\right \}. \label{eq3}
\end{align}
Equation \ref{eq3} can be implemented by softmax self-attention, for details, see Appendix~\ref{sec:A sfmx}.

Our main departure here from standard kernel gradient descent is variable step size $\eta(X)$, which we term \textit{context-adaptive learning rate}. An important property is that $\eta(X)$ depends not only on the points in the dataset/context, but also on the position of $\bx_\q$ relative to them. If there are many points near the query exemplar, $\eta(X)$ will be relatively \textit{smaller}, evoking some of the flavor of non-parametric algorithms.

Notably, the two ``effective parameters'' $c_\sigma, c_\eta$ in the softmax case (in combination with the context-adaptive learning rate) differ from the single ``effective parameter'' in the linear case (learning rate $\eta$), pointing to how transformers using these two different activation functions may learn differently.

\section{Transformers \textit{do} learn kernel gradient descent to solve classification tasks}
\label{sec:experiments}
Knowing theoretically that self-attention is expressive enough to do an (adaptive) step of (kernel) gradient descent in the classification setup, the key question becomes if this solution is present in trained transformers, i.e. does a trained transformer actually implement GD on the data in context? In this section, we present empirical evidence towards a positive answer by training transformers using the setup from Section~\ref{sec:setup}. To show the similarity between the two algorithms we make use of two different approaches. All implementation details can be found in Appendix~\ref{sec:A impl}.

\textbf{Metrics similarity per context sample.} Firstly, we observe per-sequence metrics in both models. Concretely, metrics we observe are loss, entropy, and probability of the correct class. For each of the metrics, we plot $N=512$ points representing different contexts: $x$-axis being the metric value (of predicting $\by_\q$) obtained from a trained transformer's forward pass, and the $y$-axis is metric value obtained with a step of gradient descent.
The indicator two algorithms are similar is all the points concentrating around the $y=x$ line.

\textbf{Models alignment.} For completeness, following the alignment metrics introduced by \citep{von2023transformers}, we investigate the similarity of the two models by capturing three quantities throughout training, for each averaging the values over $N=100$ contexts:
\begin{itemize}
    \item Prediction differences (Preds diff) - the norm of the differences in the predicted outputs by the two models ${\|\hat \by_{\q}^{\text{TR}} - \hat \by_{\q}^{\text{GD}}\|}$
    \item Cosine similarity (Cos sim) - the cosine similarity between the sensitivities of the two models $\frac {\partial (\hat \by_\q^{\text{TR}})_j} {\partial \bx_\q}$ and $ \frac {\partial (\hat \by_\q^{\text{GD}})_j} {\partial \bx_\q}$, averaged over classes $j=1,...,C$
    \item Sensitivity differences (Model diff) - The norm of the differences in sensitivities of the two models $\frac 1 C \sum_{j=1}^C \|\frac {\partial (\hat \by_\q^{\text{TR}})_j} {\partial \bx_\q}- \frac {\partial (\hat \by_\q^{\text{GD}})_j} {\partial \bx_\q}\|$
\end{itemize}
\subsection{Linear transformer learns gradient descent}
To test whether trained linear SA actually learns the solution provided in Proposition \ref{prop1}, we compare its performance with one step gradient descent (starting from $W=\bold 0$), whose learning rate value is found through a linear search. In Figure~\ref{fig:lin5}, we show the results for $d=5, C=5, n=100$. For different values of $d$, see Appendix~\ref{sec:A exp lin}. The low values of the prediction and sensitivity differences, the cosine similarity converging to 1, and the similar per-sample metrics suggest that a single-layer linear transformer does learn to solve linear classification task by doing one step of GD.
\begin{figure}
    \centering
    \includegraphics[width=\linewidth]{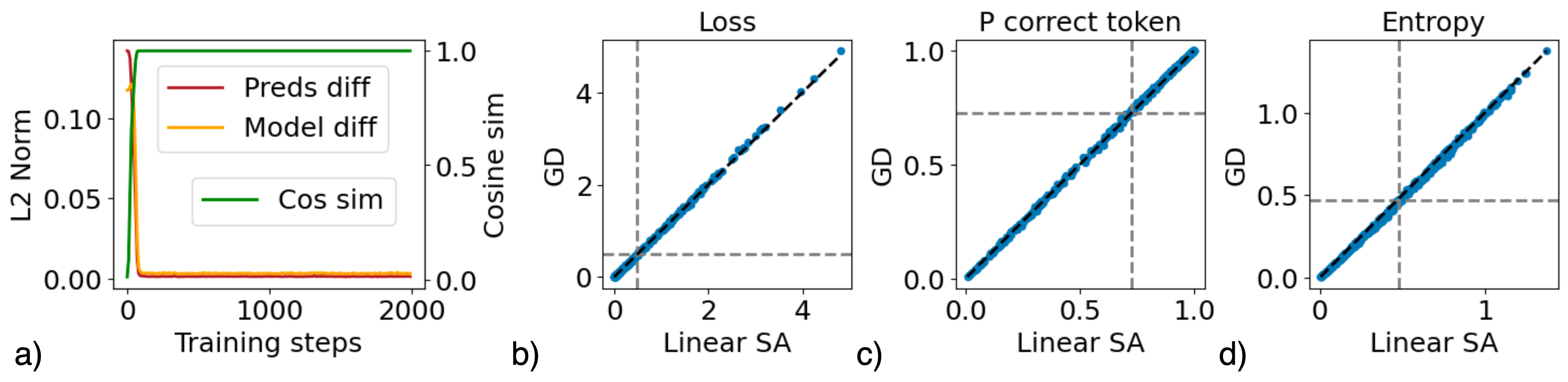}
    \caption{Similarity between the two algorithms---trained linear SA and a GD step---in the setup with $C=5, n=100, d=5$. a) Alignment metrics through transformer training. \textit{Right:} Similarity per context sample of b) loss, c) probability of the correct class and d) entropy of $\hat \by_\q$: each point represents the value on one context. Dotted line corresponds to the mean value of a metric.}
    \label{fig:lin5}
    \vspace{-0.3em}
\end{figure}
\subsection{Softmax transformer (mostly) learns context-adaptive kernel gradient descent}
\label{sec: softmax does impl}
\begin{figure}
    \centering
    \includegraphics[width=\linewidth]{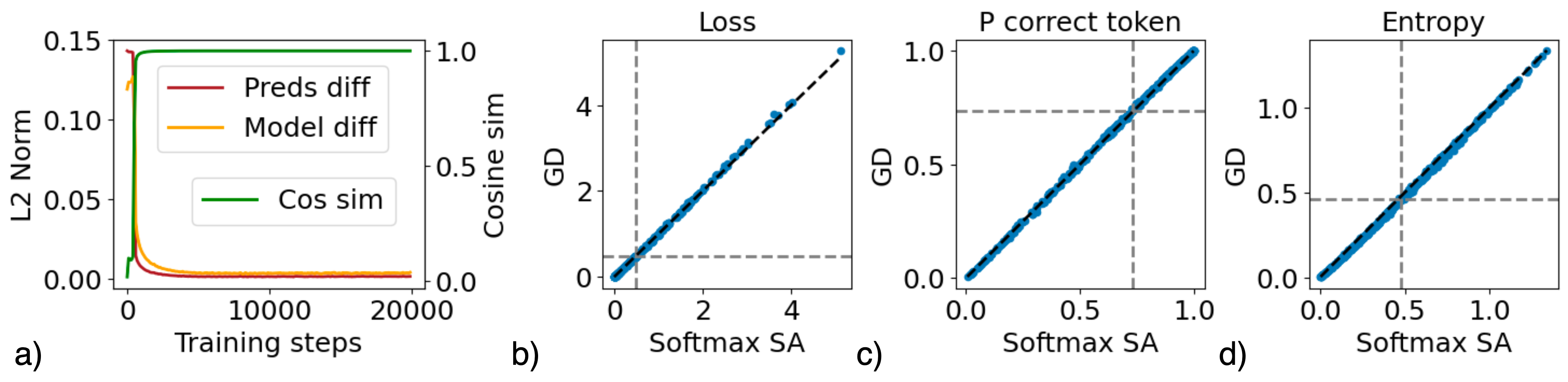}
    \caption{Similarity between a trained softmax SA and a context-adaptive step of kernel GD, in the setup $C=5, n=100, d=5$. a) Models alignment metrics through transformer training. \textit{Right:} Metrics similarity per context sample: b) loss, c) probability of the correct class and d) entropy achieved by both algorithms.}
    \label{fig:s5}
    \vspace{-0.5em}
\end{figure}

Following the interpretation of softmax SA implementing one context-adaptive step of GD in the kernel space (Section~\ref{sec:theory_sfmx}), we show that such a solution can be present in a trained transformer. For the vanilla context-adaptive step of kernel GD (Equation \ref{eq3}), we pick $c_\eta$ and $c_\sigma$ that minimize CE loss in a grid-search. We compare this to a softmax SA trained on our toy classification task contexts. The comparison of the two algorithms is again indicating high similarity between them. In Figure~\ref{fig:s5}, we observe that softmax SA gets very similar to one context-adaptive step of kernel GD, and that per context sample prediction metrics are almost equal. The experiment here uses $d=5, C=5, n=100$; for different setups, see Appendix~\ref{sec:A exp sfmx}.
\paragraph{Selecting vs eliminating: softmax attention does both.} When we train softmax transformers on the linear classification task, we observe a surprising bifurcation in behaviors learned. We term them ‘selection' and ‘elimination'. In selection (the more common strategy), the transformer attends to similar exemplars in-context and copies forward the corresponding label. In elimination, the transformer instead attends to the most \textit{distant} exemplars and eliminates the corresponding labels. We further analyze the latter behavior in Appendix~\ref{sec:A sel vs el}, and focus the main paper results on `selection'.
\paragraph{Dynamics and meta-learning the kernel shape.} As demonstrated empirically, a transformer learns to implement Equation~\ref{eq3}, corresponding to a single step of GD in kernel space with a context-adaptive learning rate. This suggests that a trained softmax SA---with parameters $W_Q, W_K, W_V, W_O$---can be effectively characterized by just two parameters: $c_\eta$ and $c_\sigma$. Observing that the learned weights $W_Q^\top W_K$ and $W_V^\top W_O^\top$ resemble those from our construction in Appendix~\ref{sec:A sfmx}, we found a method to extract the effective values of $c_\sigma$ and $c_\eta$ from the trained model. More details on this, as well as observed dynamics of these parameter evolutions are discussed in Appendix~\ref{sec: A extract}). Here, we note the key takeaway---for the different setups (in our case varying the dimension $d$), softmax SA leverages its ability to adapt the kernel width to the task at hand---it effectively meta-learns the optimal width.

\section{Implications: Softmax attention is more expressive than linear on classification tasks}
\label{sec:softmax>linear}
In this section, we aim to compare ICL learning algorithms---the one learned by linear SA, and the one learned by softmax SA. The theory from Section~\ref{sec:theory} guides our intuition. A transformer's training process on our task can be viewed as a form of meta learning. Our results show that linear SA performs a step of GD on the classification loss, while softmax SA performs a step of kernel GD (for the Gaussian RBF kernel) with a context-adaptive learning rate.  The linear case has a single effective parameter ($\eta$), while the softmax case introduces two ($c_{\sigma}$, $c_{\eta}$), potentially improving adaptation to complex task distributions. Furthermore, GD implemented by the softmax SA features a context-adaptive learning rate $\eta(X)$, which may also help with extracting more signal from context.

\begin{figure}
    \centering
    \includegraphics[width=0.95\linewidth]{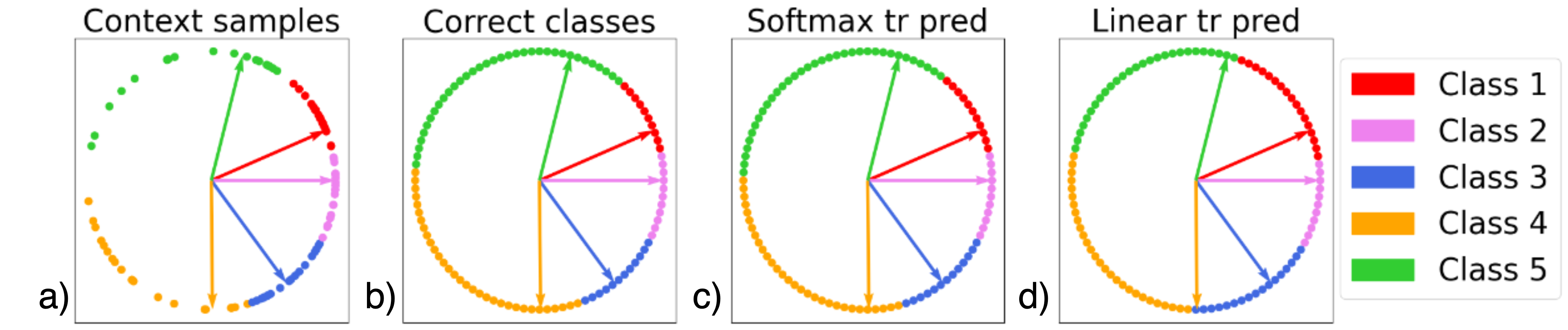}
    \caption{The qualitative differences between softmax and linear ICL algorithms, plotted by fixing a context and attaching different points around the circle as queries. a)~The fixed context we use. b)~Ground truth labels algorithms should predict. c)~The predictions made by softmax transformer. d)~The predictions of linear transformer. We notice how linear transformer struggles with classes 1 (red) and 3 (blue), as they occupy smaller, denser regions of the circle.}
    \label{fig:diffs2d}
    \vspace{-0.5em}
\end{figure}

To qualitatively investigate the differences, we plotted the decision boundaries of the two algorithms in the setting of $d=2, C=5, n=100$; see Figure~\ref{fig:diffs2d}. We observe that the linear SA may have a problem ``focusing'': due to linear attention, even if the query point is given in the context, it happens that the attention paid to it is outnumbered by many relatively close-by points of a different label.

Strictly speaking, there are two reasons why softmax transformer may perform better: (1) ability to meta-learn the kernel width $\sigma^2$ and (2) use of context-adaptive learning rate. We decouple the two properties by (conceptually) ablating each. First, we consider a step of kernel gradient descent---kernel GD (Equation \ref{eq2}), with RBF kernel, where the hyper-parameters $\eta, \sigma^2$ are found in grid-search (thus ablating the context-adaptive learning rate). Second, we train a softmax self-attention, but keeping the $W_Q^\top W_K$ matrix fixed to the construction from Appendix~\ref{sec:A sfmx}, with $c_{\sigma}=\sqrt{d+C}$. This effectively results in a kernel width fixed to $\sigma^2=1$, i.e. fixing the ``shape'' of the attention---similar to the fixed attention shape in linear SA. Note that the point of the latter run is to test whether having only adaptive learning rate can lead to the same level performance as the softmax transformer, hence we don't fix $\sigma^2$ to a near-optimal value. The summary of the models and properties they satisfy can be found in Table \ref{tab:4models}.
\begin{table}[h]
  \caption{Four models we compare and their features}
  \label{tab:4models}
  \centering
  \begin{tabular}{lcc}
    \toprule
    \textbf{Model} & \textbf{Optimal kernel width} & \textbf{Context-adaptive $\eta(X)$} \\
    \midrule
    Linear SA & \xmark & \xmark \\
    Softmax SA, $\sigma^2=1$ & \xmark & \cmark \\
    Kernel GD & \cmark & \xmark \\
    Softmax SA & \cmark & \cmark \\
    \bottomrule
  \end{tabular}
\end{table}
\begin{figure}[!ht]
    \centering
    \includegraphics[width=0.77\linewidth]{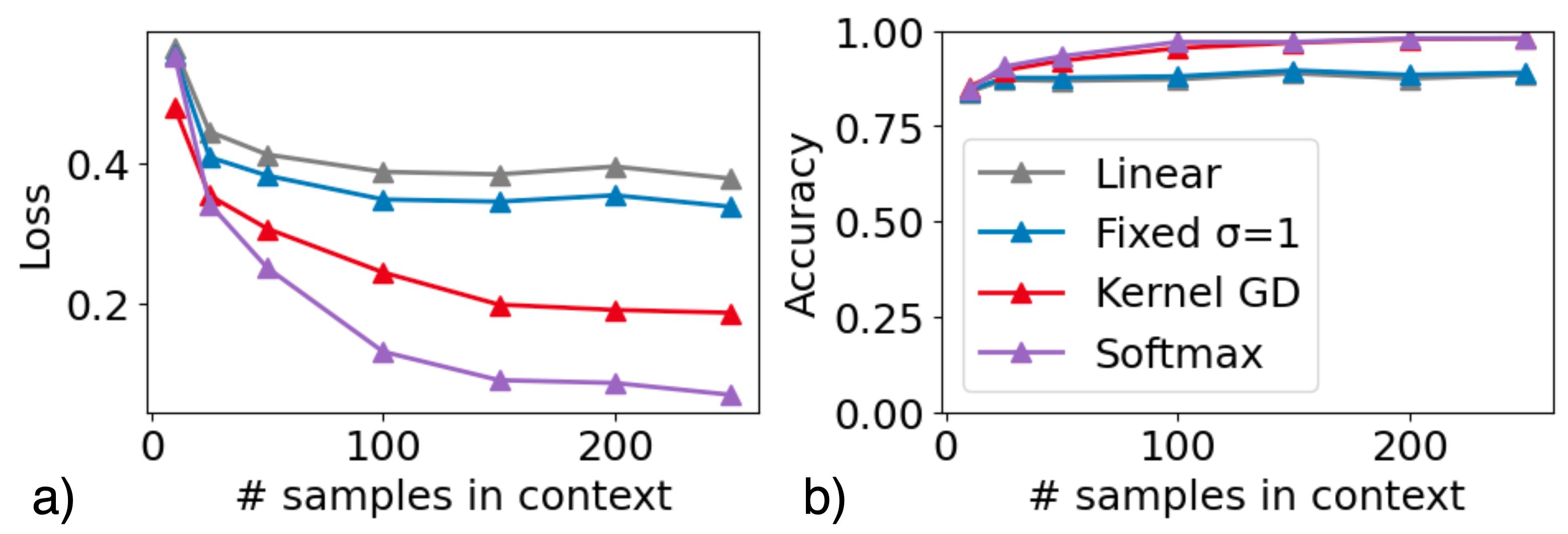}
    \caption{a) Loss and b) accuracy of the predictions that the four models from Section~\ref{sec:softmax>linear} give, plotted as a function of context length.}
    \label{fig:4alg}
    \vspace{-0.5em}
\end{figure}

The metrics achieved by the four models can be seen in Figure~\ref{fig:4alg}, plotted as the context length increases. From the metrics of softmax SA with a fixed kernel width, we see that to a fixed, but not near-optimal kernel shape, adaptive learning rate doesn't help too much. Together with the discussion in Section~\ref{sec: softmax does impl}, this demonstrates one of the key advantages of softmax attention over linear---the ability to meta-learn the correct kernel shape. Kernel GD on its own is highly accurate (essentially matching the performance of full softmax SA), but lacks confidence/higher saturation in its answers, as we can infer from higher loss. Increasing the learning rate $\eta$ would lead to higher confidence in predictions (see Equation \ref{eq2}), but this might actually lead to worse overall loss due to high confidence on incorrect answers.\footnote{Indeed, CE loss averaging over one highly saturated correct prediction, and a highly saturated incorrect prediction is very large (due to the latter), unlike that of averaging a near-uniform-distribution correct and incorrect prediction.} Recall that learning rate for kernel GD is found by a grid-search that optimizes the CE loss, so it balances these factors, leading to worse loss overall. This points to the intuition that the gap between a trained softmax transformer and the grid-search fitted kernel GD is due to the context-adaptive learning rate---it leads to higher confidence on "easier" contexts, so the softmax transformer is rarely too confident \textit{and} wrong. We further verify this empirically in Appendix~\ref{sec:A adaptiveness}, by decoupling "easy" and "hard" settings and the way softmax transformer deals with them. The key takeaway of experiments in this section is that softmax and linear SA are different---softmax SA benefits both from the meta-learned kernel width and the context-adaptive learning rate.

\section{Conclusion}
The remarkable ICL ability of transformers is a key reason behind their success in generalizing to never-before-seen sequences, both in controlled small-scale settings like ours and at a much larger scales of models such as LLMs. Approaching ICL from a theoretical perspective, we demonstrate that softmax self-attention is expressive enough to implement a single context-adaptive step of kernel gradient descent. Furthermore, we design a synthetic classification task and use it to show that softmax self-attention learns to implement this theoretical solution in practice. Exploring the differences between the ICL algorithms implemented by linear and softmax SA, we conclude that both the ability to meta-learn the kernel width and the context-adaptive learning rate are the key factors that enable softmax transformers to outperform their linear counterparts on the challenging settings of our task.

Prior theoretical work on ICL \cite{garg2022can, von2023transformers} demonstrated that self-attention can implement a step of gradient descent, but under simplifying assumptions---specifically, using linear attention and MSE loss on a regression task. We bridge these gaps to more practical transformer settings by investigating softmax self-attention on a classification task with CE loss. To connect to empirical ICL results, we reproduce the phenomenon of ICL transience \cite{singh2023transient} in our setting (Appendix~\ref{sec:A transience}), providing further evidence that it is a simple but meaningful and worthwhile framework to study.

Altogether, we hope this paper lays theoretical foundations for exploring ICL in settings that are closer to practical applications. While important challenges remain---such as understanding multi-layer architectures and the role of MLPs---we believe progress will come by systematically relaxing simplifying assumptions, one step at a time.
Foundational understanding is not only important for theoretical clarity, but also essential for driving meaningful, lasting progress.

\section*{Acknowledgements}
We thank Basile Confavreaux, Jin Hwa Lee, Yedi Zhang, Stephanie C. Y. Chan and Andrew K. Lampinen for useful discussions. This
work has been supported by the Gatsby Charitable Foundation (GAT3850).

\bibliographystyle{unsrtnat}


\appendix
\section{Extended Related Work}
\label{sec:related_work}
The concept of emergent in-context learning was first observed in \cite{brown2020language}, where it was noted that the trained GPT-3 language model could perform well on few-shot tasks presented in the prompt, without any parameter updates. Since then, ICL has been compared to meta-learning \cite{andrychowicz2016learning,finn2017model} and so-called mesa-learning \cite{hubinger2019risks}, where trained models are interpreted as performing optimization at inference time.

Staying close to practice, there has been extensive empirical analysis of ICL, with several works investigating how well LLMs perform in in-context learning settings, as well as their capabilities and limitations \cite{liu2021makes, zhang2022opt, wei2023larger, bhattamishra2023understanding, li2024long, zhao2024probing}. To further investigate the phenomenon, researchers have begun designing carefully crafted experimental setups, usually involving smaller transformers and non-language datasets, such as Omniglot \cite{chan2022data, singh2024needs}, boolean function classes \cite{bhattamishra2023understanding}, modular arithmetic \cite{he2024learning} or Markov chains \cite{edelman2024evolution, park2024competition}. An important component of transformers' ability to perform ICL has been attributed to the formation of so-called induction heads \cite{olsson2022context, reddy2023mechanistic, singh2024needs, edelman2024evolution}. Additionally, ICL has been shown to be a transient phenomenon \cite{singh2023transient, anand2024dual, chan2024toward, yin2025attention, carroll2025dynamics, singh2025strategy}, where, if a task can be solved through both in-context and in-weights learning, the model eventually transitions to solving it via in-weights learning—that is, by memorizing the example-label pairs in its parameters. In this case, data properties also play a significant role in the emergence and persistence of ICL \cite{chan2022data}.

From a more theoretical perspective, prior work has typically focused on small transformers (1–2 layers), often combined with additional simplifications for mathematical convenience. The prior work referenced in this paragraph adopts the controlled ICL setting, where the context consists of input-output pairs drawn from a task—each context corresponding to a different task.

\textbf{Linear self-attention.} Another commonly used simplification of self-attention is the variant without the softmax activation, often referred to as linear self-attention. Linear self-attention is expressive enough to solve linear regression task by performing one step of GD, and it also does learn it when trained on ICL linear regression (each context corresponds to different linear regression task), as shown by prior work \cite{von2023transformers, garg2022can, akyurek2022learning, ahn2023transformers, mahankali2023one}. The mathematical simplicity of linear attention has facilitated further theoretical research on ICL, revealing additional properties and learning dynamics \cite{wu2023many, zhang2024trained, lu2024asymptotic, vladymyrov2024linear, zhang2024context,zhang2025training}.

\textbf{Self-attention with non-linear activation.} Due to the analytical complexity, there haven't been as many theoretical results on attention mechanisms with non-linearities. \citet{ahn2023transformers, bai2023transformers} have provided expressivity results (weight constructions) for ReLU-activated self-attention, demonstrating its ability to implement various algorithms for solving various tasks. Softmax attention has been explored in combination with continuous tasks, using MSE loss \cite{cheng2023transformers, huang2023context, collins2024context, chen2024training, yang2024context, wang2024context}. \citet{collins2024context} shows that softmax self-attention adapts to function Lipschitzness; however, this result isn’t applicable in our setting, as classification tasks aren’t continuous (i.e. have infinite Lipschitzness). \citet{cheng2023transformers} mention the similarity between softmax transformer and a step of gradient descent in the kernel space; we carefully examine this statement and explore the cases of our classification setup where the two algorithms perform differently. Moving closer to the matter in this paper, there have been studies on in-context binary \textit{classification} using softmax self-attention, but together with hinge loss \cite{li2024nonlinear}, and MSE loss \cite{li2024one}. Our goal is to take a step further to understanding softmax self-attention in the ICL classification setting with \textit{cross-entropy} loss, as this is the setup used with transformers in practice (e.g., in LLMs). \citet{wang2024understanding} do explore transformers with CE loss on a distinct classification task, and show how they approximate kernel GD step for finite dimensional kernels. After extending the proof to the infinite dimensional spaces, we further investigate the differences between this approximation and the actual trained transformer via observing the impact of the context-adaptive learning rate.

\section{Advantages of the context-adaptive learning rate}
\label{sec:A adaptiveness}
Continuing the discussion in Section~\ref{sec:softmax>linear}, we experimentally investigate the differences between a step of kernel gradient descent, with and without the context-adaptive learning rate, as well as the benefits it brings.

\begin{figure}
    \centering
    \includegraphics[width=\linewidth]{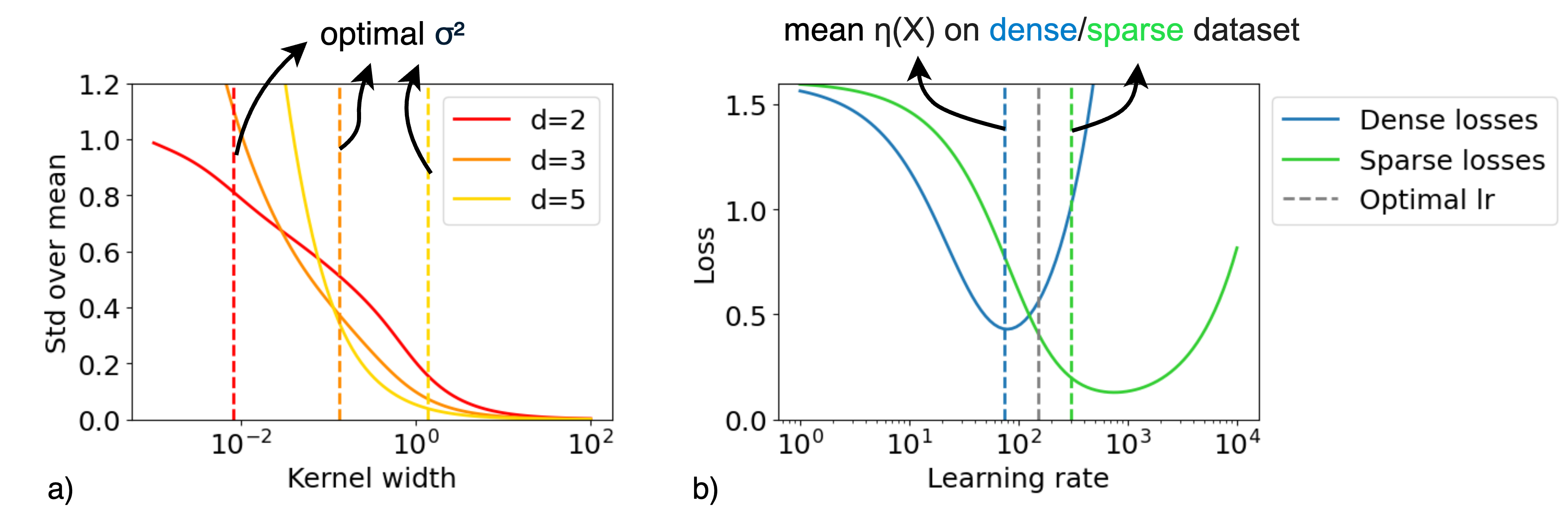}
    \caption{a) Standard deviation over mean of the quantity from Appendix~\ref{sec:A adaptiveness} (that we are approximating by a constant), as a function of kernel width $\sigma^2$. The grid-search optimal values of $\sigma^2$ are denoted with dotted lines. This explains the differences in a step of kernel GD with and without the adaptive learning rate. b) Loss dependence on the learning rate in two setups---$\bx_\q$ being in a dense or sparse region. The optimal learning rates differ, which can't be captured by a constant $\eta$---so it has to compromise between the two (gray dotted line). However, when allowing for adaptive learning rate $\eta(X)$, kernel GD can capture the difference---as denoted by accordingly colored dotted lines.}
    \label{fig:adaptive}
\end{figure}

\paragraph{Variance of the context-adaptive learning rate $\eta(X)$.} For the fixed number of points in context $n=100$ and $C=5$, the largest difference happens in small dimensions, such as $d=2,3$. Namely, for such small $d$, the optimal kernel width (found by grid search) is smaller, implying that it is optimal to cast attention only to tokens that are much closer to $\bx_\q$. This makes sense, because in smaller dimensions, the context points are over-all closer to each other, as we could see in Figure~\ref{fig:task_examples}. However, this makes it harder to approximate the context-adaptive learning rate $\eta(X)=\frac {c_{\eta}e^{1/\sigma^2}n} {\sum_{i=1}^ne^{\bx_i^\top \bx_{\q}/\sigma^2}}$ with a single constant $\eta$. We empirically show the root of the problem, plotting the standard deviation over mean of the quantity $\frac {\sum_{i=1}^ne^{\bx_i^\top \bx_{\q}/\sigma^2}} {e^{1/\sigma^2}}$ for $N=100$ different contexts. The results, shown in Figure~\ref{fig:adaptive}a), indicate that when the dimension is low, the optimal kernel width lays in the region of high variance (relative to the mean) of the quantity we're approximating. 

\paragraph{Decoupling two optimal modes for learning rate value.} The cause of high variance in the sum $\sum_{i=1}^ne^{\bx_i^\top \bx_{\q}/\sigma^2}$ is related to the way we sample points in our task. Because each class has equal number of samples in context, when more classes are closer together, they create a \textit{dense region} of points on the sphere. In contrast, when a region isn't highly populated, due to only one or two class vectors nearby, the points will be further apart, creating a \textit{sparse region} of points. Contexts without highly dense/sparse regions are ``easier" for kernel GD without the adaptive learning rate, in the sense that one learning rate fits all contexts better. To illustrate the trouble, for $d=2, n=100, C=5$ we take $N=3996$ contexts with sparse and dense regions, and put the query point precisely in the most dense, respectively sparse, region. With these dense and sparse datasets, we plot the loss achieved as a function of learning rate in kernel GD (for a fixed, generally optimal $\sigma^2$). The results, shown in Figure~\ref{fig:adaptive}b, depict the difference in the optimal learning rate, depending on whether the query point is located in the dense or the sparse region of the context. Adaptive learning can account for this difference, as seen from the colored dotted lines, making advantage over the fixed learning rate (gray dotted line). Further explanations of the generation process for this experiment can be found in the Appendix~\ref{sec:A impl}.

 \paragraph{Context-adaptive learning rate adapts to task difficulty.} Apart from the differences in optimal learning rates depending on whether $\bx_\q$ is in dense or sparse region (Figure~\ref{fig:adaptive} b), we can see the differences in the loss---indicating that it is easier to predict the correct label when $\bx_\q$ is in sparser region. This makes sense intuitively as well, because in sparse regions there is very likely only one dominating class. The adaptive learning rate $\eta(X) = \frac {c_{\eta}e^{1/\sigma^2}n} {\sum_{i=1}^ne^{\bx_i^\top \bx_{\q}/\sigma^2}}$ accounts exactly for that---it corrects the confidence based on how hard the task is, depending on the position of $\bx_\q$. Recall that higher confidence means that the model predictions are more saturated, which is achieved by higher $\eta$. For context-dependent learning rate, $\eta(X)$ is lower (i.e. confidence is lower) when $\sum_{i=1}^ne^{\bx_i^\top \bx_{\q}/\sigma^2}$ is higher, meaning there are many points in the neighborhood of $\bx_\q$ which likely indicates more than one class vector around. On the other hand, $\eta(X)$ gets higher when $\sum_{i=1}^ne^{\bx_i^\top \bx_{\q}/\sigma^2}$ is lower and not many points are close by---meaning the task has tendency to be easier, so model should be more confident in its answer.

\section{Transience of ICL in our setting}
\label{sec:A transience}
We approached the question of understanding ICL from theoretical perspective, and that is the main focus of this paper. Nonetheless, we explore the connection with a different approach: trying to understand ICL from more practical, empirical point of view. One of the results discovered in this way is the transient nature of ICL \cite{singh2023transient, park2024competition, chan2024toward, singh2025strategy}---if the model has an option to solve a task both by using the context (ICL), and by memorizing (in its weights) all the input-output mappings (called in-weights learning (IWL)), it will first do ICL, but gradually switch to IWL. In this section, we aim to reproduce this phenomena in our toy classification setting. Successfully observing this transition within our controlled setup would strengthen the validity of our setting and theoretical assumptions.
\subsection{Setup for ICL transience}
In the generation process we used so far, each context has its own set of class vectors determining how the class labels are being assigned---if we train on 200000 batches of 2048 contexts, this results in $200000\times 2048$ different sets of class vectors through training, which is not possible to be memorized in model's weights. To get around this, we introduce $m$ (16 in our example below) different sets of class vectors, and through training, we provide contexts labeled according to these only, each batch containing an equal number of contexts labeled according to each set of class vectors.

The way transience is usually observed is by looking at the accuracy through training of the model on different evaluation contexts---the ones that can be answered only if the model uses ICL, i.e. information from the context rather than the stored input-output maps. In our setting, that translates to using context with randomly generated class vectors for evaluation (as we have been using so far for theory part). We use the setting of $C=5, d=3, n=30$ (6 samples per class in one context). 

\begin{figure}[!ht]
    \centering
    \includegraphics[width=\linewidth]{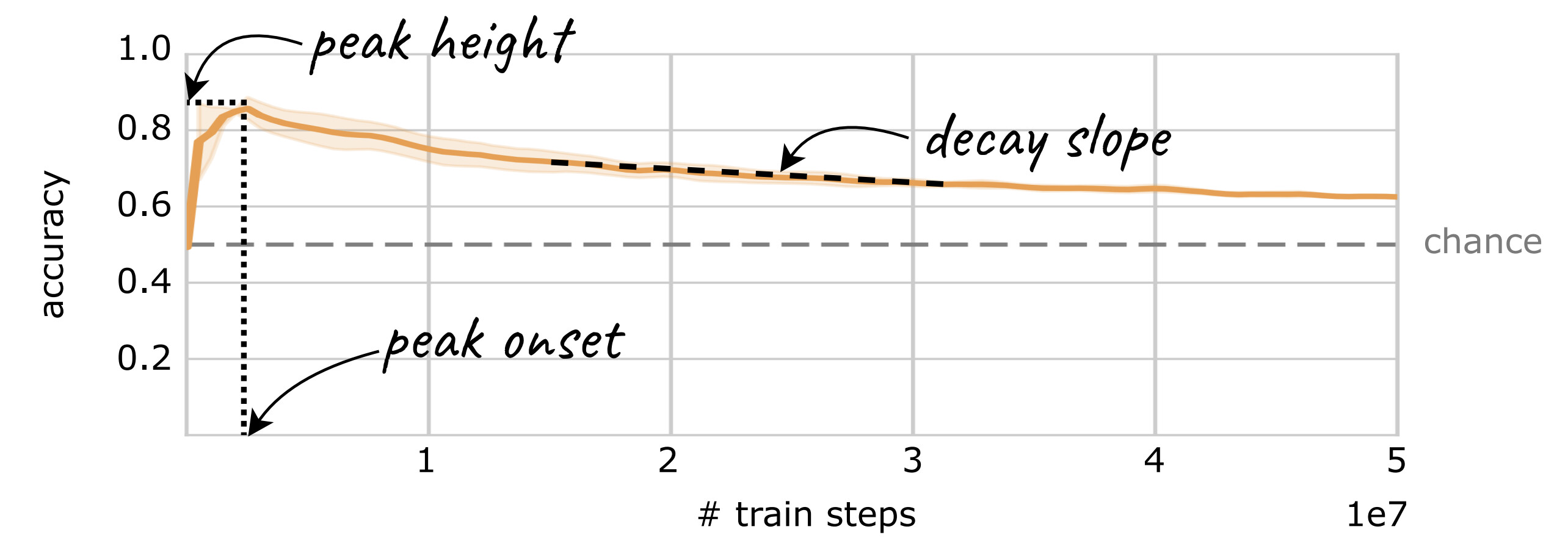}
    \caption{Reproduced, with permission, from \citet{singh2023transient}: depicts the evaluation accuracy of ICL task through training, where training task can be solved by both ICL and IWL. The drop in the accuracy after the peak indicates ICL mechanism slowly fading from the model.}
    \label{fig:tr_a}
\end{figure}

Figure~\ref{fig:tr_a} is from the paper \cite{singh2023transient} that first observed transience in a (completely different from ours) binary classification setup---it shows the accuracy on ICL evaluation contexts, while the model is training on the contexts allowing both ICL and IWL mechanisms. We see that initially, the model scores high accuracy on ICL evaluation, before continuing on to decrease with a slow decay slope, indicating slow disappearing of the ICL mechanism through training. 
\subsection{Transience in our setting}
\begin{figure}
    \centering
    \includegraphics[width=\linewidth]{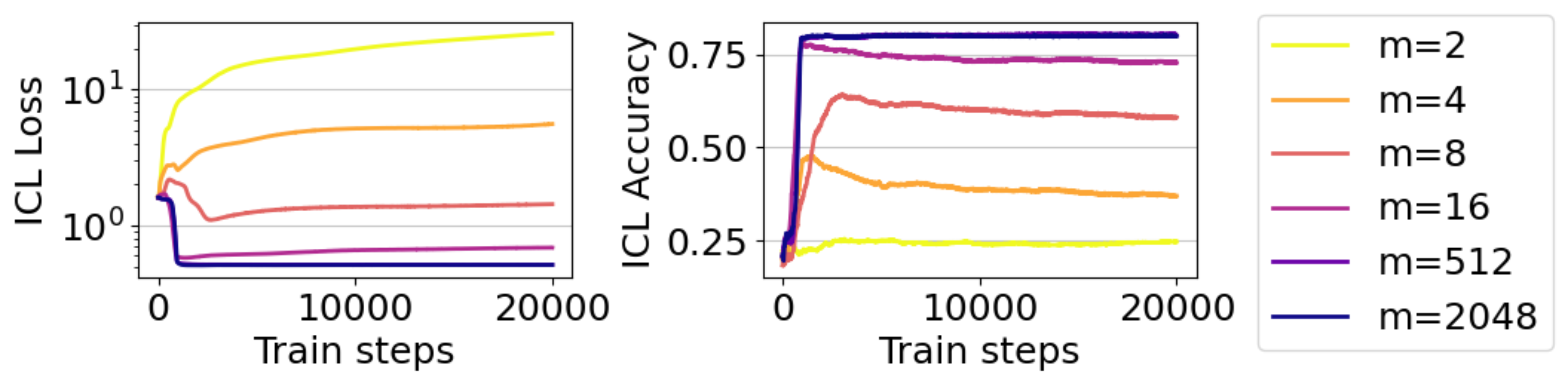}
    \caption{ICL loss (left) and ICL accuracy (right) when the model is trained with contexts labeled by only $m$ different sets of class vectors. For the different values of $m$ the model uses different strategies (ICL vs IWL); interestingly, ICL strategy fades with training in some cases ($m=4,8,16$).}
    \label{fig:1l_tr}
\end{figure}

Following prior work, we use softmax self-attention. To bring our setup a step closer to practical transformers, and enable the model to perform in-weights learning, we now introduce a multi-layer perceptron (MLP) after the self-attention layer. In order to observe a full spectrum of behaviors, we plot the ICL evaluation loss and accuracy for different values of $m$. The results in Figure~\ref{fig:1l_tr} show the variety of task solutions the model learns to implement. For smaller values, such as $m=2$, transformer seems to immediately find the IWL solution, hence the large ICL loss and low ICL accuracy. As $m$ grows larger, we see that ICL mechanism appears, but slowly fades away with more training. For large $m$, such as 512 and 2048, the model is more persistent with solving the task using ICL, replicating the data-dependence (with \# of classes) seen in prior work \citep{chan2022data,singh2023transient}, but in this simpler setting with fully synthetic data.

\begin{figure}
    \centering
    \includegraphics[width=\linewidth]{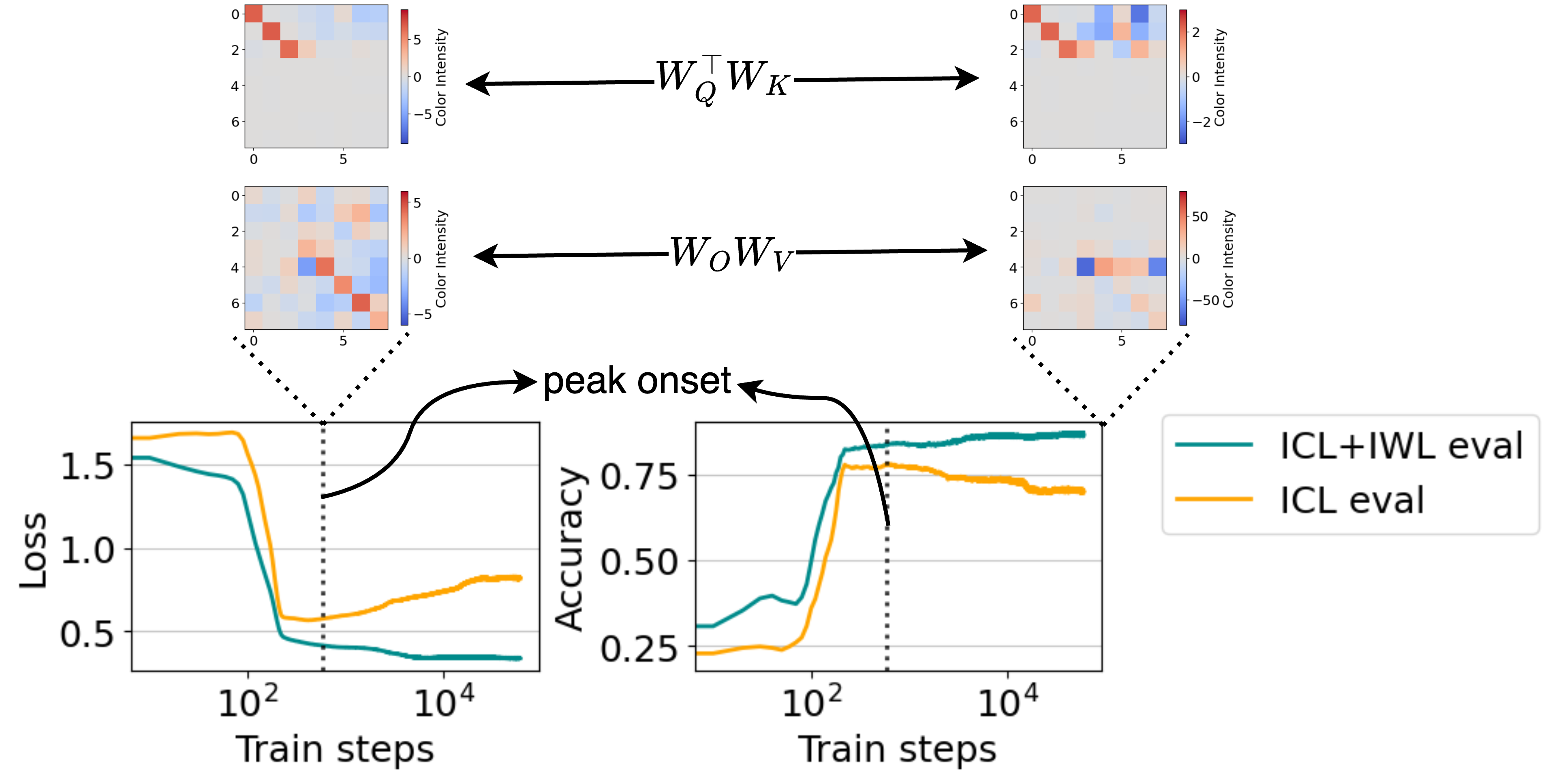}
    \caption{Loss (left) and accuracy (right) of the run trained with $m=16$ different sets of class vectors, evaluated on both ICL (random class vectors for each context) and ICL+IWL (the same $m=16$ sets of class vectors as during training) eval datasets.}
    \label{fig:icliwl_l1}
\end{figure}

In Figure~\ref{fig:icliwl_l1}, we show the loss and accuracy through training, evaluated on two different datasets: one with randomly generated sets of class vectors (ICL eval) and one where the context vectors are labeled according to only $m=16$ sets of class vectors that the model is trained on (ICL+IWL eval). We see that ICL eval has its peak, and from there accuracy is decreasing, while ICL+IWL accuracy continues to improve. This indicates that the model started focusing on learning in-weights, rather than learning to implement an ICL algorithm. To emphasize this point, we plotted model weights from the ICL peak and from the end of training; simplicity of our setup makes it possible to get insights from the weights, where prior work usually had to work around this by observing attention patterns and doing causal ablations. We observe that the weights during the ICL peak resemble the ones from our construction in Appendix~\ref{sec:A sfmx} (suggesting that the ICL algorithm in use may be context-adaptive step of kernel GD), although with some noise which likely originates from IWL mechanism starting to emerge. At the end of training, the weights changed, and don't seem to be doing GD algorithm anymore. These are harder to interpret, especially because of the potential composition happening with the MLP layer coming afterwards. For implementation details, refer to Appendix~\ref{sec:A impl}.

Even though the binary classification used by \citet{singh2023transient, singh2025strategy}, the Markov chain prediction in \citet{park2024competition} and other setups reproducing this phenomenon are completely different tasks from ours, it is interesting that transience of ICL is happening in all of them, further justifying its universality.

\section{Linear SA can do one step of GD on classification task}
\label{sec:A lin}
This section aims to provide additional justification and construct the necessary weights, thus completing the proof of Proposition \ref{prop1}, which demonstrates that a linear transformer can perform one step of gradient descent in a linear classification setting.

We first justify the derivation $\nabla_WL(\bx) = \bx(\sfmx(\bz)-\by)^\top$, where $\bz=W^\top \bx$. Let ${W=[\bw_1,..., \bw_C]}$ and $z_j=\bw_j^\top \bx$. Then $\nabla_\bz L = \sfmx(\bz)-\by$. As $\nabla_Wz_1 = [\bx, \bold 0,…, \bold 0]$, we have 
\begin{equation*}
    \nabla_W L(\bx) = \sum_{i=1}^C\frac {\partial L}{\partial z_i}\nabla_{W}z_i=\sum_{i=1}^C (p_i- y_i)[\bold 0,..., \bold 0, \bx,\bold 0,...], \text{ where }\bold p = \softmax(\bz).
\end{equation*}
Recognizing the outer product, this leads us to
\begin{equation*}
    \nabla_W L(\bx) = \bx (\softmax(\bz)-\by)^\top .
\end{equation*}

Additionally, we provide a simple set of weights that result in a transformer forward pass being the same as Equation \ref{eq1}. Namely, setting
\begin{equation*}
    W_Q^\top W_K = \begin{bmatrix}
    I_{d} & 0_{d\times C}\\ 0_{C\times d} & 0_C
    \end{bmatrix}, W_V^\top W_O^\top  = \begin{bmatrix}
    0_{d} & 0_{d\times C}\\0_{C\times d} & \frac \eta NI_C
 \end{bmatrix}
\end{equation*}
enables extracting $\bx_i$s from the merged tokens in the attention, and $\by_i$s in the value and projection matrix.
\section{Kernel activation SA can do one step of kernel GD on classification task}
Here, we provide the missing part of the proof of Proposition \ref{prop2}, namely that Equation \ref{eq2} holds for all kernels $k$, as well as the transformer weights implementing it. All together, it shows that a kernel activation transformer can implement one step of gradient descent in the kernel space.
\subsection{Functional gradient descent leads to Equation \ref{eq2}}
\label{sec: A ker theory}
Let $X=\R^d$ be the space of inputs, and $k:X\times X\to\R$ a kernel. Let $\mathcal H$ be the reproducing kernel Hilbert space (RKHS) associated with $k$. We do a functional gradient descent in $\mathcal H$, making use of chain rule. Corresponding to the $W_0=\bold 0$ assumption of Proposition \ref{prop2}, we start from $f_1,...,f_C\in \mathcal H$ being the zero functions, where $\softmax(f_j(\bx))_{j=1}^C$ gives the predicted probabilities of $\bx$ being of classes $j=1,...,C$. Now, we lay out similar equations as in the case of normal gradient descent:\footnotetext{I.e. $E_\bx(f) = f(\bx)$, for all $f\in \mathcal H$.}
\begin{align*}
    L(f_1,...,f_C) &=- \by^\top \log (\softmax(\bz)) &&\text{[where $\bz=(f_1(\bx),...,f_C(\bx))^\top $]}\\
    \nabla_{f_1}L &= \sum_{i=1}^C\frac {\partial L}{\partial z_i}\nabla_{f_1}z_i &&\text{[by chain rule]}\\
    & = \frac {\partial L}{\partial z_1}\nabla_{f_1}z_1\\
    & = \frac {\partial L}{\partial z_1}\nabla_{f_1} E_\bx(f_1) &&\text{[where $E_\bx:\mathcal H\to\R$ is the evaluation functional\footnotemark]}\\
    & = \frac {\partial L}{\partial z_1} k(\bx,\cdot) &&\text{[as $\nabla E_\bx=k(\bx, \cdot)$]}\\
    & = (p_1 - y_1)k(\bx,\cdot) &&\text{[where $\bold p =\softmax(\bz)$, using $\nabla_\bz(L)=\softmax(\bz)-\by$]}
\end{align*}
where the functional gradient with respect to other functionals $f_j, j=2,...,C$ is computed similarly.
Given a sample set of points $(\bx_i, \by_i)_{i=1}^n$, the updated $f_1,...,f_C$ become
\begin{equation*}
\begin{bmatrix}
    f_{1,\text{new}} \\ f_{2,\text{new}}\\ ... \\f_{C,\text{new}}
\end{bmatrix}
=
\begin{bmatrix}
    f_{1} \\ f_{2}\\ ... \\f_{C}
\end{bmatrix} - \frac \eta n \sum_{i=1}^n(\softmax(\bz_i)-\by_i)k(\bx_i, \cdot)
\end{equation*}
Now assuming $f_i=0$, and applying these updated functions to $\bx_\q$ gives:
\begin{align*}\nonumber
    \hat \by_\q &= \softmax\begin{bmatrix}
        f_{1,\text{new}}(\bx_\q) \\ f_{2,\text{new}}(\bx_\q)\\ ... \\f_{C,\text{new}}(\bx_\q)
    \end{bmatrix}\\
    & = \softmax\bigg \{- \frac \eta n \sum_{i=1}^n (\frac 1 C\bold 1-\by_i)k(\bx_i, \bx_\q)\bigg \}\\
    & = \softmax\bigg \{ \frac \eta n \sum_{i=1}^n \by_i k(\bx_i,\bx_\q)\bigg \}.
\end{align*}
This concludes the proof that Equation \ref{eq2} holds for all kernels. $\square$
\subsection{Weight construction in Proposition \ref{prop2}}
\label{sec:A ker}
Given a self-attention whose activation function $\text{act}_k$ originates from a kernel (as described in Section~\ref{sec:model_setup}) if we set these weights:
\begin{align*}
    W_Q &= \begin{bmatrix}
    I_{d} & 0_{d\times C}\\ 0_{C\times d} & 0_C
    \end{bmatrix}
    W_K = \begin{bmatrix}
    I_{d} & 0_{d\times C}\\ 0_{C\times d} & 0_C
    \end{bmatrix}\\
    W_V &= \begin{bmatrix}
    0_{d} & 0_{d\times C}\\0_{C\times d} & I_C
    \end{bmatrix}
    W_O = \begin{bmatrix}
    0_{d} & 0_{d\times C}\\0_{C\times d} & \frac \eta NI_C
 \end{bmatrix},
\end{align*}
we recover the Equation \ref{eq2} in the transformer's forward pass.
\section{Softmax SA can do one context-adaptive step of kernel GD on classification task}
\label{sec:A sfmx}
In this section, we provide a set of self-attention weights giving raise to the Equation \ref{eq3} in the transformer's forward pass, hence showing that softmax transformer is expressive enough to implement one context-adaptive step of GD:
\begin{equation*}
    W_Q^\top W_K = \begin{bmatrix}
    c_{\sigma}I_{d} & 0_{d\times C}\\ 0_{C\times d} & 0_C
    \end{bmatrix}, W_V^\top W_O^\top  = \begin{bmatrix}
    0_{d} & 0_{d\times C}\\0_{C\times d} &c_{\eta}I_C
 \end{bmatrix}.
\end{equation*}
\section{Linear transformer does learn to do GD}
\label{sec:A exp lin}
\begin{figure}
    \centering
    \includegraphics[width=\linewidth]{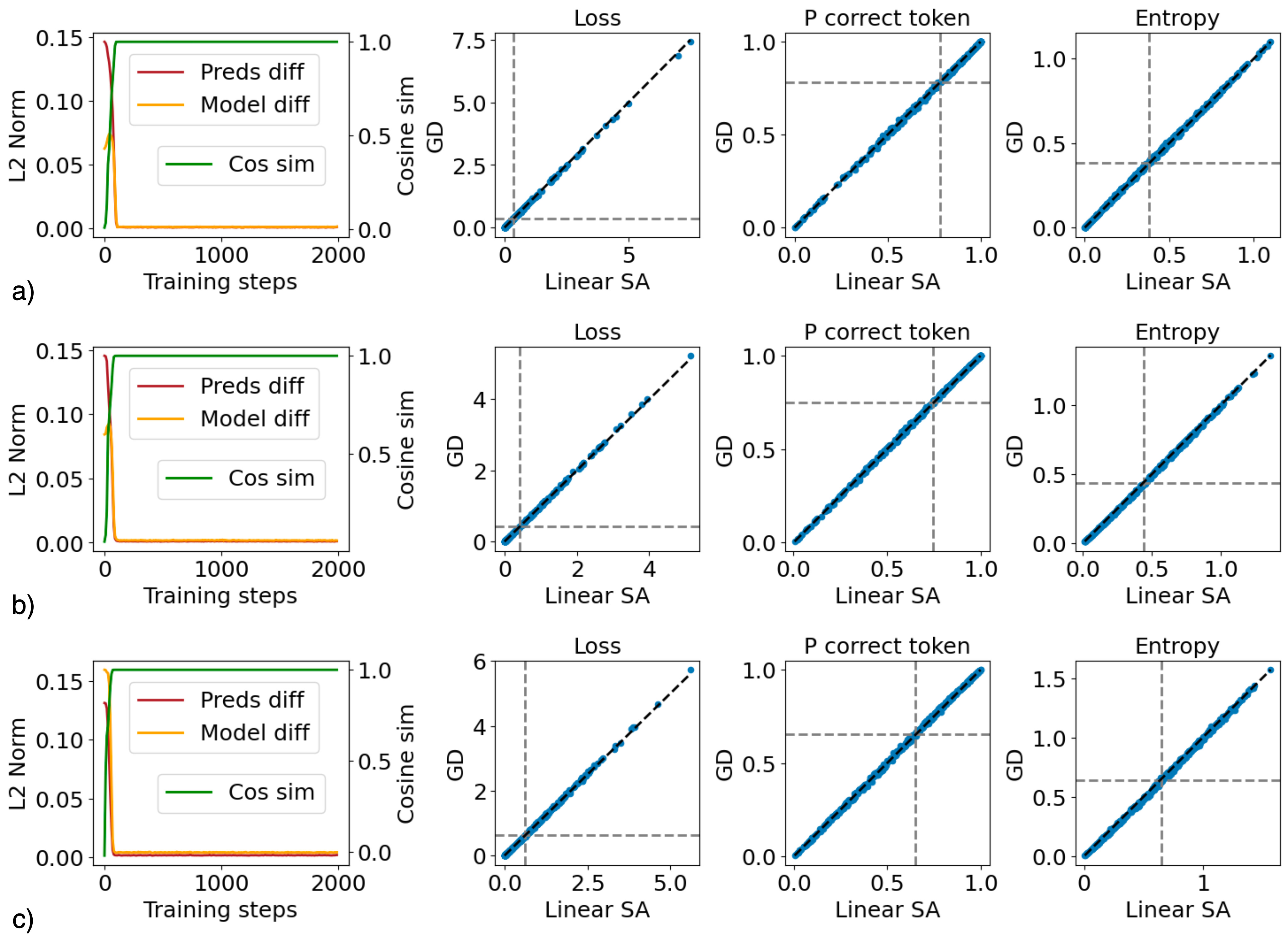}
    \caption{Shows the similarity between the two algorithms---trained linear self-attention and a GD step---in the setup with $C=5, n=100$ and: a) $d=2$, b) $d=3$, c) $d=10$.}
    \label{fig:lin}
\end{figure}

In this section, we repeat the experiments resulting in Figure~\ref{fig:lin5}, but for different dimensions of the space: $d=2,3,10$. The results are shown in Figure~\ref{fig:lin}. This is a strong argument showing that linear transformer does learn to implement one step of GD on the data it's given in the context.
\section{Softmax transformer does learn to do context-adaptive step of kernel GD}
\label{sec:A exp sfmx}
\begin{figure}
    \centering
    \includegraphics[width=\linewidth]{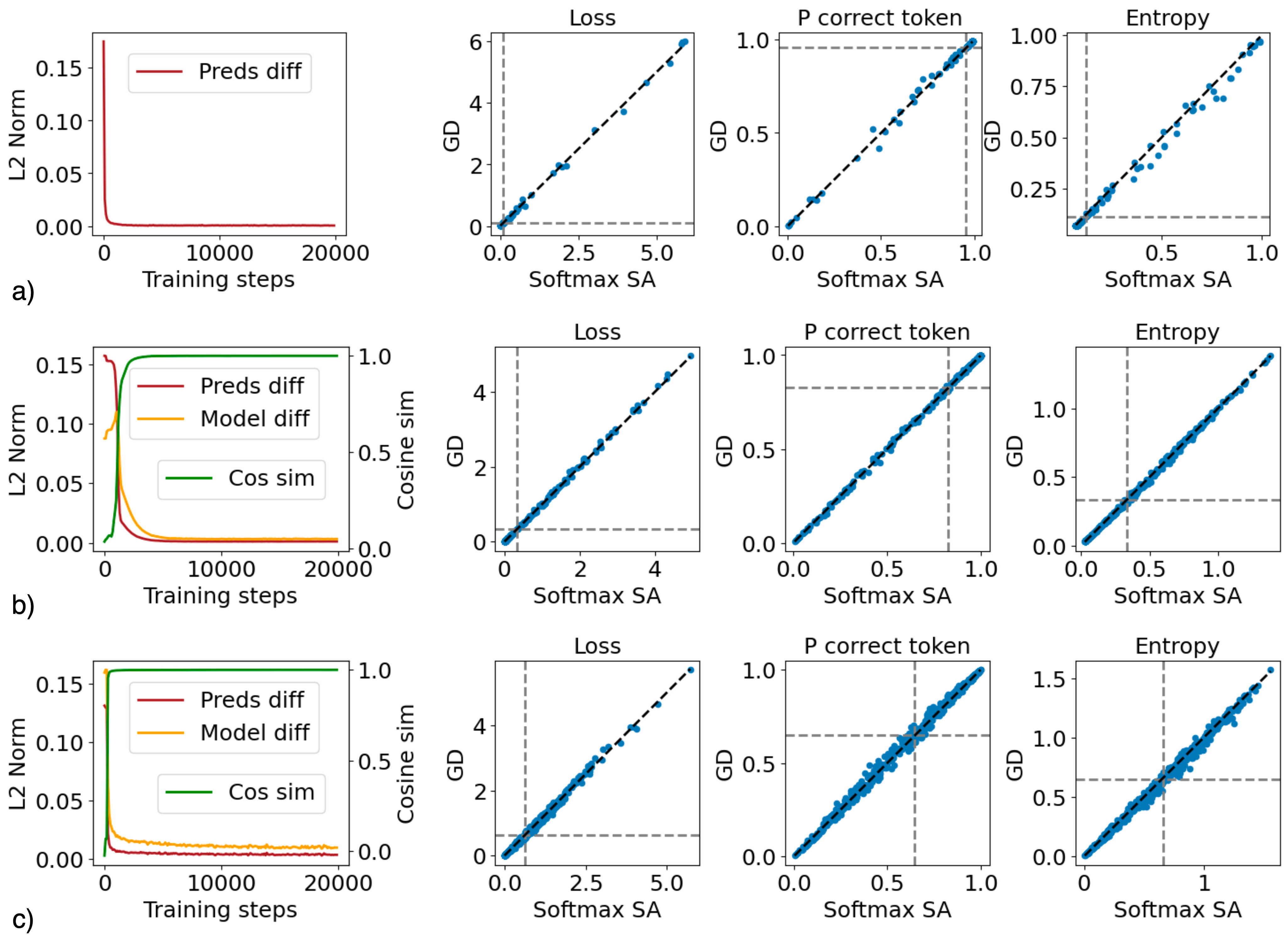}
    \caption{Similarity between a trained softmax self-attention and a context-adaptive step of kernel GD---in the setup with $C=5, n=100$ and: a) $d=2$, b) $d=3$, c) $d=10$.}
    \label{fig:sfmx}
\end{figure}

Here, we provide more results analogous to Figure~\ref{fig:s5}, varying different dimensions of the space $d$. Figure~\ref{fig:sfmx} shows the results for $d=2,3,10$. We note that for $d=2$, the gradient $\frac {\partial (\hat \by_\q^{\text{TR}})_j} {\partial \bx_\q}$ is numerically unstable to compute in our setup, so we couldn't provide the results on the cosine similarity and sensitivity differences. The numerical instability comes from highly saturated softmax attention. In the case of $d=10$, softmax attention and context-adaptive kernel GD are slightly different, and we believe that is due to the fact that the training of the transformer, in terms of the values of $c_\eta$ and $c_\sigma$, hasn't converged yet (see Section~\ref{sec: A d=10}).

\section{Implementation details}
\label{sec:A impl}
\textbf{Training.} For our empirical experiments, we train transformers in JAX \cite{jax2018github} using a modified version of the open-sourced code from \cite{singh2024needs}. The use of JAX ensures full reproducibility of our experiments, while also enabling efficient training via just-in-time compilation and automatic vectorization. We use Adam optimizer \cite{kingma2014adam} with the default parameters. All experiments were run on a single node with an NVIDIA H100 GPU. Each experiment completed in under 24 hours, with most taking significantly less time (on average ~15 minutes for linear SA and ~4h for softmax SA).

In all the experiments, we use single-headed self attention, without the causal mask. In all experiments, model training had 2048 batch size, meaning 2048 linear classification task contexts. All evaluation sets contained 512 context samples. We emphasize again that we don't use positional encoding, token embedding nor unembedding, MLPs nor layer norm\footnote{As we have only one layer, it doesn't really make a difference.}.

\textbf{Linear transformer hyper-parameters.} For linear transformer, we mostly follow the setup in \cite{von2023transformers}, with learning rate $0.00005$. On initialization, we rescale the weights by $0.002$ and perform gradient clipping with value $0.001$. We train the transformer for 200,000 iterations (batches). We evaluate every 100 iterations, hence our results are showing training steps in the range 0-2,000.

\textbf{Softmax transformer hyper-parameters.} For softmax transformer, we also rescaled the initial weights by 0.002, and we set the gradient clipping value to $1.0$. Transformer training was done over 2,000,000 iterations and (as linear) evaluated every 100 iterations, therefore shown in range 0-20,000. For fairness of comparison, Figure~\ref{fig:4alg} experiments were all done---including softmax and linear self-attention---in range 0-2,000.  Learning rate was dimension dependent: $d=2$ - $0.0006$ in Figure~\ref{fig:sfmx} and $0.006$ in Figure~\ref{fig:4alg} (as the number of training iteration was divided by 10); $d=3$ - $0.00003$; $d=5$ - $0.0001$; $d=10$ - $0.0005$.

\textbf{Gradient descent grid search.} For a GD step, we did a grid search for learning rate over $\eta\in[10^{0}, 10^{2.5}]$, trying out 100 examples chosen from a logarithmic range. For a kernel GD step, we did search for 2 hyper-parameter values, $\eta$ and $\sigma^2$. We tested 100 values of $\eta\in [10^0, 10^3]$ (log scale), times 100 values of $\sigma^2\in [10^{-3}, 10^2]$ (log scale). In both cases, the search was done using $N=10,000$ different contexts from the classification task described in $\ref{sec:task_setup}$.

\textbf{Grid search for $c_{\sigma}$ and $c_{\eta}$.} For context-adaptive kernel GD, we did grid search to find $c_{\sigma}$ and $c_{\eta}$. These are shown in Figures \ref{fig:s5} and \ref{fig:sfmx}. We used different scan ranges depending on the dimension. For $c_{\sigma}$, we tried 100 values from log range $[\text{vmin}, \text{vmax}]$, where vmin and vmax are: $d=2$ - $\text{vmin}=10^0, \text{vmax}=10^3.5$; $d=3,5$ - $\text{vmin}=10^{-1}, \text{vmax}=10^2$; $d=10$ - $\text{vmin}=10^{-3}, \text{vmax}=10^1$. For $c_{\eta}$, the 100 values were from log range with: $d=2,3,5$ - $\text{vmin}=10^{-1}, \text{vmax}=10^2$;
$d=10$ - $\text{vmin}=10^1, \text{vmax}=10^4$. To perform the search, we used $N=10,000$ different contexts from the classification task in $\ref{sec:task_setup}$.

\textbf{Softmax self-attention with fixed kernel width.} In Figure~\ref{fig:4alg}, we use softmax self-attention with kernel width $\sigma^2=1$. This was achieved by fixing the weights $W_K, W_Q$ be the ones from construction in Section~\ref{sec:A sfmx}, with corresponding $c_\sigma$. We trained for 200,000 iterations, i.e. evaluation range 1-2,000. The learning rate used is 0.0003.

\textbf{Experiments with contexts with sparse and dense regions.}
Here, we describe the generation process for the Figure~\ref{fig:adaptive}b). To generate contexts with sparse and dense regions, we sampled $N=10000$ contexts, and took $N=3996$ out of them with $\|\frac 1 n\sum_{i=1}^n\bx_i\|\ge 0.3$, with the intuition that such contexts have means of all the samples towards a particular direction, very likely indicating a sparse region and lots of classes in that direction. Those are the contexts we hypothesize contain dense and sparse regions. Next, to position the query token in the most dense, resp. sparse place, we take $K=50$ points on the circle forming a regular $K$-gon, and pick the one that has the most (resp. the least) $\bx_i$s s.t. $\bx_i^\top \bx_\q>0.3$ for the query point in context of the dense (resp. sparse) dataset. Note that these two datasets have the same $N=3996$ contexts, but the only difference is the position of $\bx_\q$ withing a context. Once the sparse and dense datasets have been generated, we fit two models on the joint dataset (union of the two): kernel GD (fitting $\eta$) and context-adaptive kernel GD (fitting $c_\eta$). In both cases we use the kernel width $\sigma^2$ found from grid-search on $N=10,000$ random contexts. For the kernel GD, we show the $\eta$ found with the dotted gray line. For context-adaptive kernel GD, using $c_\eta$ found, we compute the mean effective value of $\eta(X)$ on dense and sparse datasets respectively, shown by correspondingly colored dotted lines. 

\textbf{Transience experiments.} We train transience experiments in Figure~\ref{fig:1l_tr} for 20,000 evaluation steps (each evaluation step has 100 training iterations), with batch size 2048. The learning rate we used for all models is 0.00003. In Figure~\ref{fig:icliwl_l1}, we used 60,000 evaluation steps. For all transience experiments, transformer consists of softmax self-attention, followed by MLP with two layers, where the feature dimension between them is double the model dimension ($16=2\times 8$, the model dimension is $3+5=8$). As we're still dealing with one layer transformer, we haven't used positional nor token embeddings.

\section{Extracting $c_{\sigma}$ and $c_{\eta}$ from a trained transformer, dynamics and meta-learning the kernel width}
\label{sec: A extract}
\begin{figure}
    \centering
    \includegraphics[width=\linewidth]{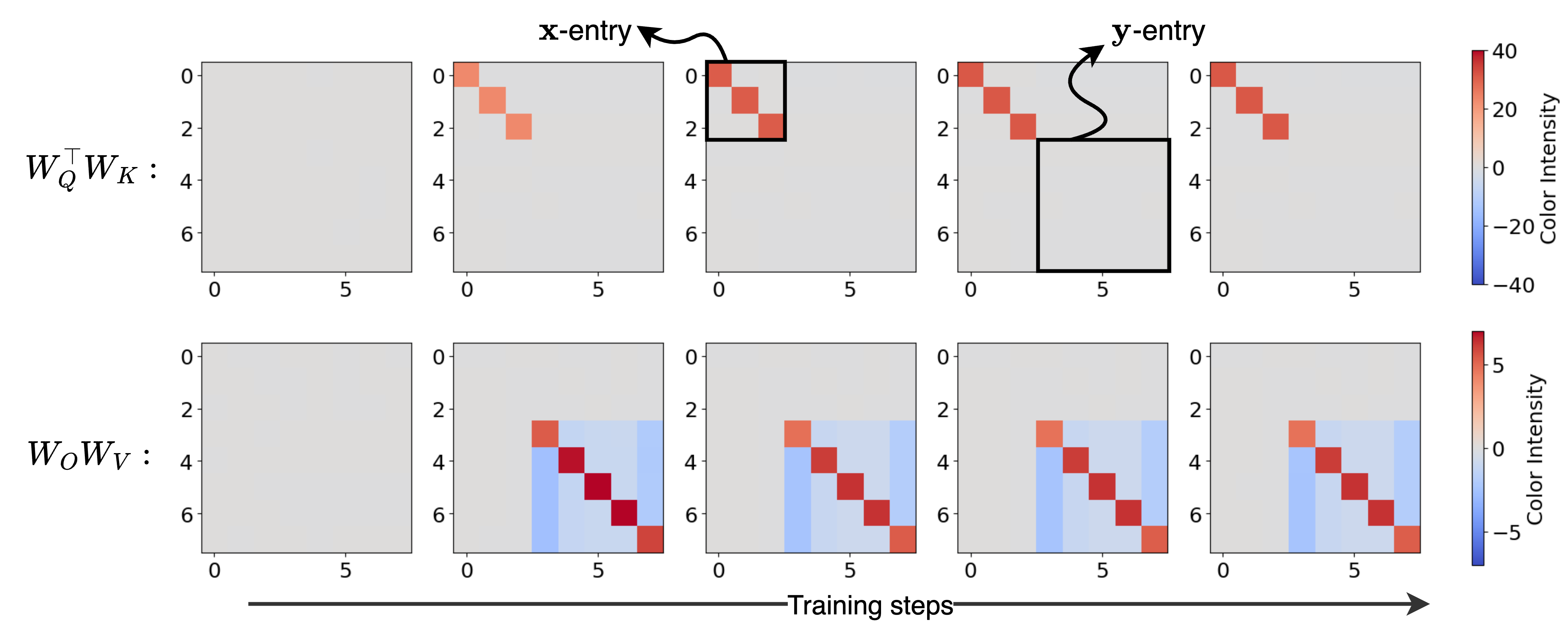}
    \caption{The weights through training: $W_Q^\top W_k$ (top row) and $W_OW_V$ (bottom row) for the setup $n=100, C=5, d=3$. Observing they look similar to the construction in Section~\ref{sec:A sfmx}, we found a way to extract the effective values $c_\eta, c_\sigma$ that softmax self-attention uses, described in Section~\ref{sec: A extract}.}
    \label{fig:weights}
\end{figure}
\begin{figure}
    \includegraphics[width=\linewidth]{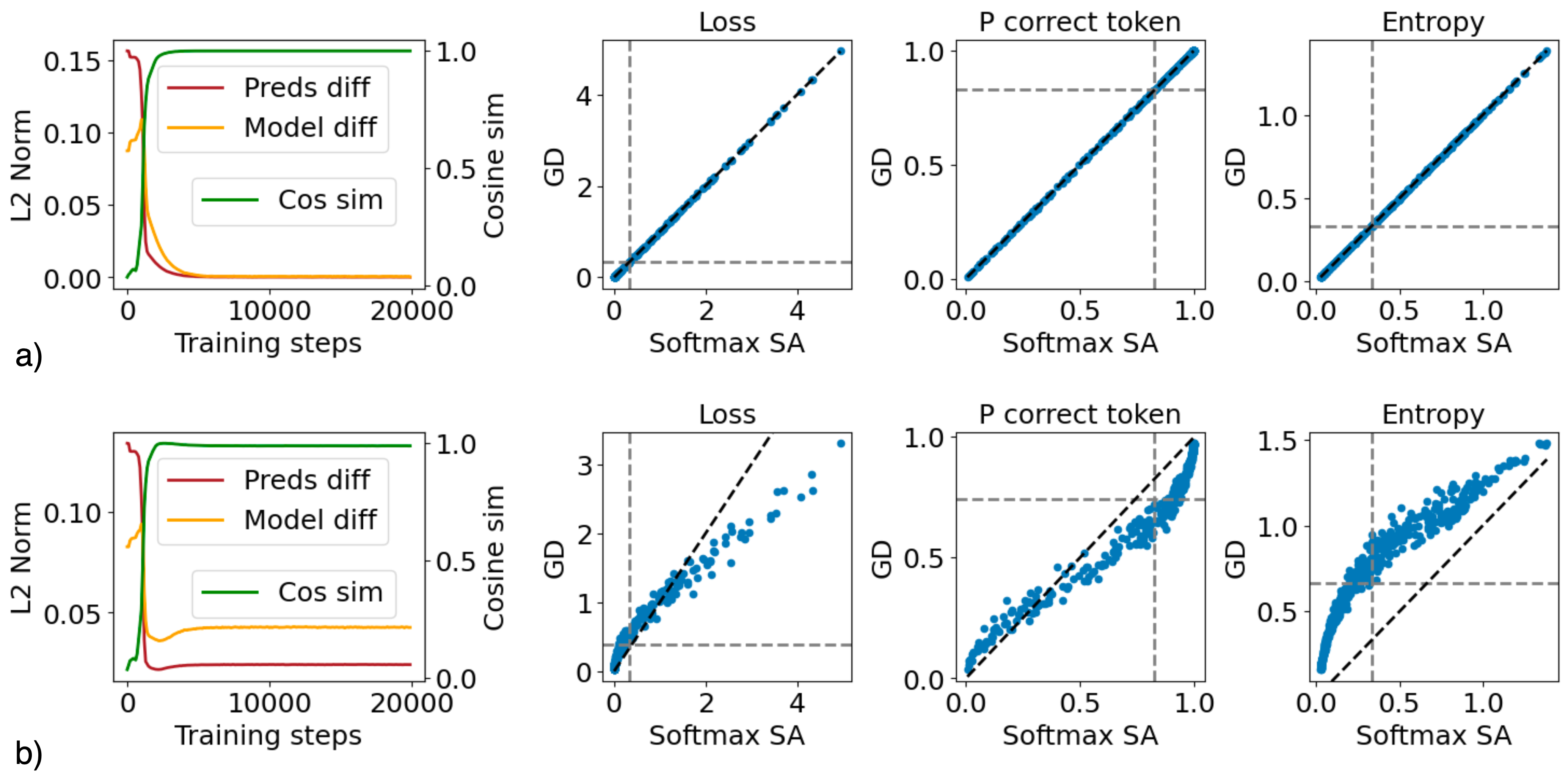}
    \caption{a) The differences between a trained self-attention and Equation \ref{eq3} with $c_\eta, c_\sigma$ values extracted with our method from the transformer. The similarities observed indicate that our method works. b) The differences between a trained self-attention and Equation \ref{eq3} with $c_\sigma, c_\eta$ being slightly off ($c_\eta=5, c_\sigma=25$) from the values extracted ($c_\eta=7.15, c_\sigma=31.07$). The observable differences further amplify the effectiveness of our method.}
    \label{fig:tr_cfit}
\end{figure}

After the theory in Section~\ref{sec:theory_sfmx}, and results from Section~\ref{sec: softmax does impl}, we concluded we can extract the effective constants $c_{\sigma}, c_{\eta}$ that softmax transformer meta-learned. Namely, if we plot the weight matrices $W_{Q}^\top W_K$ and $W_OW_V$ through training, we see they resemble the construction in the Appendix~\ref{sec:A sfmx} (see Figure~\ref{fig:weights} for the case $n=100, C=5, d=3$). To extract the effective $c_{\sigma}$, we take the average value at the diagonal of $\bx$-entry submatrix of $W_Q^\top W_K$. Extracting $c_{\eta}$ goes as follows: we first note that adding a constant to a column in $\by$-entry submatrix of $W_OW_V$ mathematically doesn't change anything---due to softmax activation being shift invariant. Thus to get to the form of our construction in Appendix~\ref{sec:A sfmx}, we add to each column of $\by$-entry the average value of the off-diagonal entries in that column. This results in the off-diagonal entries being approximately 0, so we take the average of the diagonal entries to be the effective value $c_{\eta}$. Our experiments show that this method indeed extracts the correct effective values---in Figure~\ref{fig:tr_cfit} we can see that in the setting $n=100, C=5$ and $d=3$, transformer's similarity metrics align well with the implementation of the Equation $\ref{eq3}$ with the effective values of $c_{\eta}, c_{\sigma}$ we extracted. For reference, we also included an example with $c_\eta, c_\sigma$ slightly off from the extracted values---we see that differences are large, strengthening correctness of our extraction method.

\begin{figure}
    \centering
    \includegraphics[width=\linewidth]{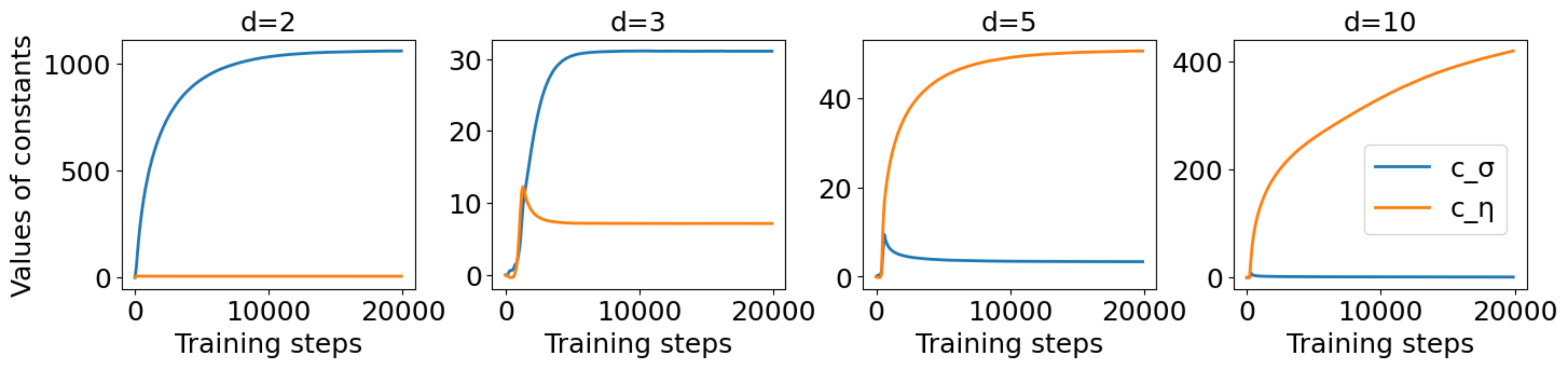}
    \caption{Dynamical evolution of $c_\sigma$ and $c_\eta$ through softmax self-attention training process. The figure depicts the setup $C=5, n=100$ and $d$ being 2,3,5 and 10, from left to right.}
    \label{fig:dyn}
\end{figure}

Figure~\ref{fig:dyn} shows the how the two effective values evolve through training, for fixed $n=100, C=5$, and varying $d$. General dynamics involves the initial spike---the time that weights of the transformer align to do kernel GD, followed by scaling of $c_\eta$ and $c_\sigma$. For $d=2$ softmax transformer learns to pay attention only to points very close to the query point (high $c_{\sigma}$). However, this isn't the case in general, as we can see for higher dimensional spaces. As the dimension gets higher, less and less points are in a close neighborhood of $\bx_\q$, with many more being approximately orthogonal to it. This leads to larger attention window (made possible with lower $c_{\sigma}$), and higher saturation/confidence in the samples that are captured in that window (higher $c_{\eta}$). In the case of $d=10$, we see that the values haven't converged yet---see Appendix~\ref{sec: A d=10} for more experiments and discussions on this setup. Key takeaway is that softmax transformer leverages its ability to adapt the kernel width to the task at hand---it effectively meta-learns the optimal width.
\section{Softmax attention selecting vs eliminating}
\label{sec:A sel vs el}
\begin{figure}
    \centering
    \includegraphics[width=0.9\linewidth]{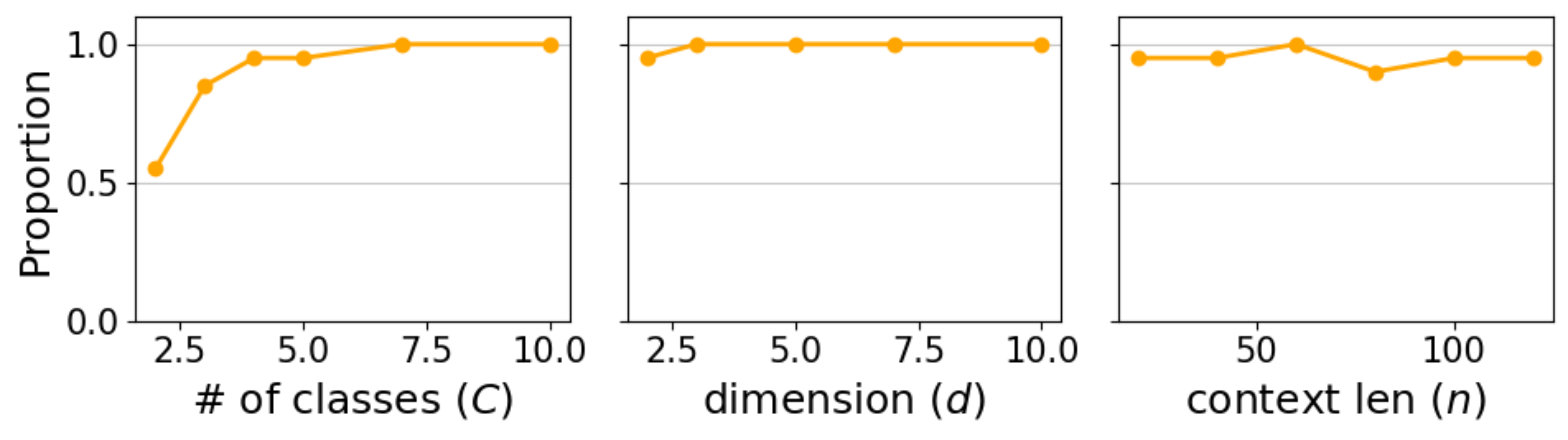}
    \caption{Proportion of runs where selection algorithm is learned, over 20 random seeds, varying number of classes, dimension and context length, from left to right.}
    \label{fig:sel_vs_el}
\end{figure}

Softmax transformer learns (at least) two different algorithms through training. In the more common case, our results match the context-adaptive kernel GD solution (Figures~\ref{fig:s5}, \ref{fig:sfmx}). This strategy can be viewed as predicting query class logits as a weighted average of $\by_i$s, weighted based on how close $\bx_{\q}$ is to different $\bx_{i}$s. Intuitively, the model is \textit{selecting} the context points from which to copy the labels, similar to the mechanism of induction heads \cite{olsson2022context,reddy2023mechanistic,singh2024needs}, with the radius of selection controlled by the kernel width $\sigma^2>0$. More rarely, we observe a different algorithm---\textit{elimination}---whereby softmax self-attention learns to attend to the tokens \textit{furthest} from the query and then \textit{subtract} those out.\footnote{Such a mechanism was more frequent when $C$ was small, indicating possible connections to the anti-induction heads noted by \cite{singh2024needs}.} Throughout this paper, we focused on the selection algorithm, because it is the one learned in majority of cases. However, especially when $C$ is low, the elimination algorithm is appearing as well. In Figure~\ref{fig:sel_vs_el}, we provide the proportion of times the selection algorithm is learned, across different seeds (20 different combinations in total). The base setting is $d=2, C=5, n=100$. For each of the plots, we make number of classes, dimension and context length, respectively, vary and observe how does the percentage of selection algorithm learned change. These experiments affirm that selection algorithm is the one leaned in majority of runs, especially in our default settings. We leave further studies of the elimination algorithm and the conditions it appears in for future work.

\section{Finding $c_{\eta}$ and $c_{\sigma}$ in the case $d=10$}
\begin{figure}
    \centering
    \includegraphics[width=0.9\linewidth]{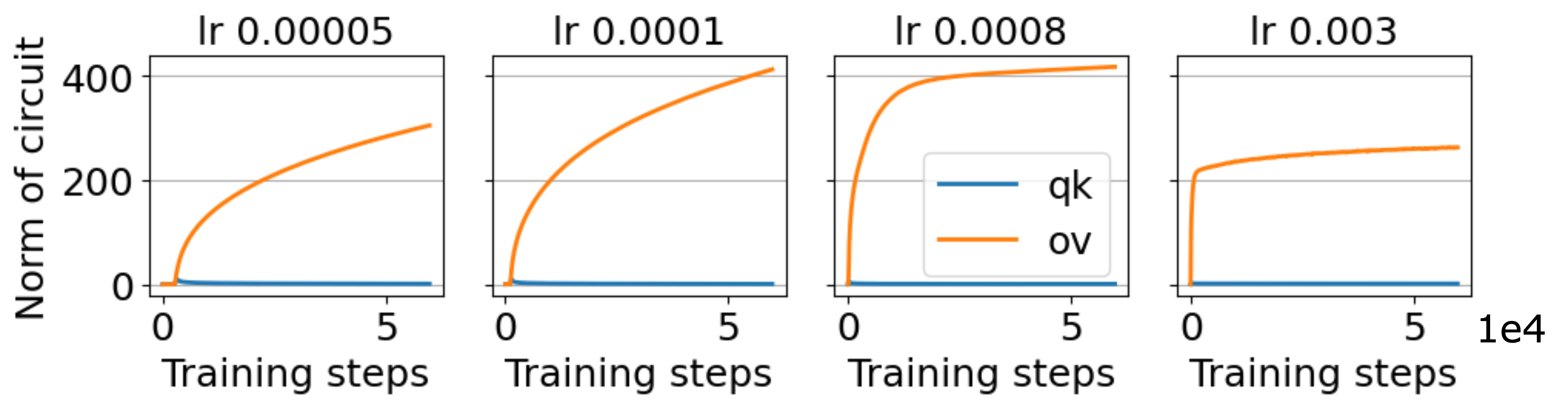}
    \caption{Evolution of the effective values of the constants $c_\eta$ and $c_\sigma$ through training, for different values of the (transformer's training) learning rate: $0.00005, 0.0001, 0.0008$ and $0.003$, respectively.}
    \label{fig:lr10}
\end{figure}

After plotting the effective value evolution for the case $d=10$ (and $C=5, n=100$), we've noticed that the growth of $c_\eta$ hasn't finished yet. Trying out different (model training) learning rates, we plot the dynamics over 60,000 evaluation iterations, seen in Figure~\ref{fig:lr10}. Interestingly, for higher learning rate, the growth of $c_\eta$ flattens earlier and around a smaller value, as seen from the right two plots. Observation that no value of $c_\eta$ converged yet, together with the fact that grid search finds the optimal $c_\eta$ to be $\sim$8,000, we realized that this case of $d=10$ has a more complicated loss landscape. This evidence suggests that a learning rate scheduler may be needed to fully successfully train a softmax SA in this case. Since this is not the primary focus of the paper, as well as our compute power restriction, we leave further investigations for future work. A lesson learned from these experiments is that seemingly flat loss and accuracy curves can be misleading, and the model may still be developing.
\label{sec: A d=10}

\end{document}